\pdfoutput=1

\documentclass[11pt]{article}

\usepackage[table]{xcolor}

\usepackage[preprint]{acl}

\usepackage{times}
\usepackage{latexsym}

\usepackage[T1]{fontenc}

\usepackage[utf8]{inputenc}

\usepackage{microtype}

\usepackage{inconsolata}

\usepackage{graphicx}

\usepackage{booktabs}
\usepackage{array}
\usepackage{multirow}
\usepackage{amsfonts,amssymb}
\usepackage{bm}
\usepackage{amsmath}
\usepackage{microtype}
\usepackage{bm}
\usepackage{pifont}
\usepackage{makecell}
\usepackage{graphicx}
\usepackage{booktabs}
\usepackage{subcaption}
\usepackage{caption}
\usepackage{float} 
\usepackage{adjustbox}
\usepackage{longtable}
\usepackage{amsmath}

\usepackage{fontawesome5}
\title{A Rose by Any Other Name: LLM-Generated Explanations Are Good \\Proxies for Human Explanations to Collect Label Distributions on NLI}

\author{
 \textbf{Beiduo Chen\textsuperscript{\faMountain\kern1pt\faRobot}} \quad
 \textbf{Siyao Peng\textsuperscript{\faMountain\kern1pt\faRobot}} \quad
 \textbf{Anna Korhonen\textsuperscript{\faSchool}} \quad
 \textbf{Barbara Plank\textsuperscript{\faMountain\kern1pt\faRobot}}
\\
\textsuperscript{\faMountain} MaiNLP, Center for Information and Language Processing, LMU Munich, Germany \\
\textsuperscript{\faRobot} Munich Center for Machine Learning (MCML), Munich, Germany \\
\textsuperscript{\faSchool} Language Technology Lab, University of Cambridge, United Kingdom \\
\tt{ \href{mailto:beiduo.chen@lmu.de}{\textcolor{black}{beiduo.chen@lmu.de}},
\href{mailto:siyao.peng@lmu.de}{\textcolor{black}{siyao.peng@lmu.de}}, 
\href{mailto:alk23@cam.ac.uk}{\textcolor{black}{alk23@cam.ac.uk}},
\href{mailto:b.plank@lmu.de}{\textcolor{black}{b.plank@lmu.de}}
}}

\begin{document}
\maketitle
\begin{abstract}

Disagreement in human labeling is ubiquitous, and can be captured in human judgment distributions (HJDs). Recent research has shown that \textit{explanations} provide valuable information for understanding human label variation (HLV) and large language models (LLMs) can approximate HJD from a few human-provided label-explanation pairs. However, collecting explanations for every label is still time-consuming. This paper examines \textit{whether LLMs can be used to replace humans in generating explanations for approximating HJD.}
Specifically, we use LLMs as annotators to generate model explanations for a few given human labels. We test ways to obtain and combine these label-explanations with the goal to approximate human judgment distributions. We further compare the resulting human with model-generated explanations, and test automatic and human explanation selection.
Our experiments show that LLM explanations are promising for NLI: to estimate HJDs, generated explanations yield comparable results to human's when provided with human labels. Importantly, our results generalize from datasets with human explanations to i) datasets where they are not available and ii) challenging out-of-distribution test sets.

\end{abstract}

\section{Introduction}

Human judgment distribution (HJD, \citealt{DBLP:journals/tacl/PavlickK19,DBLP:conf/emnlp/NieZB20,DBLP:conf/emnlp/Chen0PLKP24}) refers to the distribution of labels assigned to a specific instance by a large group of human annotators, capturing human label variation (HLV, \citealt{plank-2022-problem}). It provides rich information related to uncertainty and plausible multi-choices that should not be discarded as noise~\cite[e.g.][]{DBLP:journals/aim/AroyoW15,plank-etal-2014-learning, DBLP:journals/jair/UmaFHPPP21}.
For example, concerning the same premise-hypothesis pair in the Natural Language Inference (NLI, \citealt{DBLP:conf/mlcw/DaganGM05,DBLP:conf/emnlp/BowmanAPM15,DBLP:conf/naacl/WilliamsNB18, Manning2006LOCALTI}), different coders may perceive the relationship differently.

\begin{figure}[t]

        \centering
        \includegraphics[width=\linewidth]{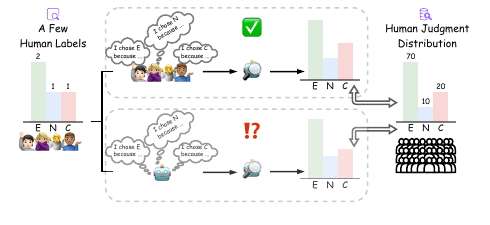}
        \caption{Recent research has shown that LLMs can approximate a human judgment distribution (HJD) in natural language inference (NLI) with the help of human explanations, as shown in the upper part. While human explanations are still relatively expensive and scarce in most datasets, we ask in the lower part: \textit{Can LLMs provide reasonable generated explanations for different NLI labels to approximate HJD?} 
        }
        \label{fig:overall}
\end{figure}

Recent research proposed using Large Language Models (LLMs) as annotators to reduce annotation cost~\cite{DBLP:conf/emnlp/TanLWBJBKL0024,DBLP:conf/naacl/HeLGJZLJYDC24}. Some works directly solicited human judgment distribution~\cite{DBLP:conf/emnlp/LeeAT23,madaan2024lost} with mixed results~\cite{pavlovic2024understanding}. In contrast, \citet{DBLP:conf/emnlp/Chen0PLKP24} used a few labels and \textit{human-provided explanations} from \citealt{DBLP:conf/acl/Weber-GenzelPMP24}) to help  Llama~\cite{DBLP:journals/corr/abs-2407-21783} and Mixtral~\cite{DBLP:journals/corr/abs-2401-04088} to effectively approximate HJD in ChaosNLI, the latter crowd-sourced 100 annotations for each NLI instance, establishing a relatively stable HJD \citep{DBLP:conf/emnlp/NieZB20}. 
They find that the resulting LLM-based model judgment distributions (MJDs) closely align with HJD. While this approach avoids the need for large-scale human annotations, it still requires human-given explanations that are far more costly to obtain than annotations of NLI labels alone.


Recent studies have found that LLMs can effectively provide explanations for tasks such as reasoning, sentiment analysis, and even business processes~\cite{DBLP:journals/corr/abs-2210-06726,DBLP:journals/corr/abs-2310-11207,DBLP:journals/corr/abs-2401-12846}. We instead study automatic explanation generation for the NLI task.
In this paper, we investigate if LLMs could provide reasonable explanations for NLI instance labels,  
and whether the generated explanations are of sufficient quality to replace costly human-provided explanations. 
Our key idea is shown in Figure~\ref{fig:overall}.

Specifically, we let LLMs generate model explanations supporting each NLI label, respectively.
We first examine ($\mathrm{i}$) can LLMs approximate HJD via diverse model explanations without access to label annotations?
The result is positive but does still fall short of the performance of LLMs with human explanations. 
We then consider using a few human labels to guide the selection of model explanations to help generate MJDs, and ask ($\mathrm{ii}$) are model explanations comparable to human's when helping LLMs to approximate HJD?
Experiments show that
MJDs from \textit{LLMs and model explanations} result in comparable scores with MJDs from \textit{LLM and human explanations}
---``A rose by any other name would smell as sweet.''\footnote{A quote from Romeo and Juliet used to metaphorically argue the intrinsic qualities or nature of something remain the same, regardless of its name or origin. }
Furthermore, we ask ($\mathrm{iii}$) does our approach extend to NLI datasets that do not have human-provided explanations? We extend the proposed method to \textit{datasets without explanations}, showing our method generalizes to this more common scenario.

Our findings are:\footnote{LLM-generated and human-validated explanations are publicly available at \href{https://github.com/mainlp/MJD-Estimator}{https://github.com/mainlp/MJD-Estimator} for reproduction.}
\begin{itemize}
    \item Model explanations are comparable to human explanations in approximating HJD on NLI, and can be scaled up from a few annotations of datasets without explanations.
    \item Results on the out-of-domain ANLI dataset show that modeling HLV information can improve NLI classifiers' performance.
    \item A human annotation study and ablation shows that explanation variability may serve as a potential indicator for evaluating HLV, and the relevancy of explanations is crucial.
\end{itemize}

\newcolumntype{P}[1]{>{\ttfamily\raggedright\arraybackslash}p{#1}}

\section{Generating Model Explanation}
\label{sec:model-ex-generate}

\begin{figure}[t]

        \centering
        \includegraphics[width=\linewidth]{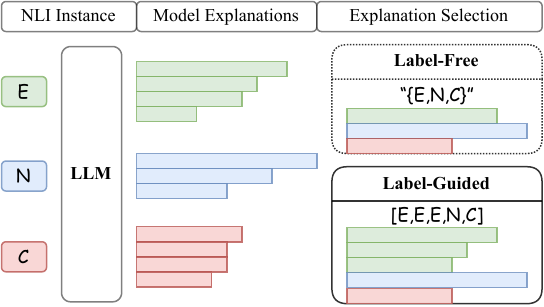}
        \caption{Illustration of the process of generating model explanation using the \textit{longest} explanation selection heuristic. 
        The \textit{label-free} scenario uniformly selects one explanation per \{E, N, C\} label, whereas the \textit{label-guided} approach follows human NLI labels to select three longest E, one longest N, and one longest C.}
        \label{fig:autoex-generation}
\end{figure}

Collecting human explanations for an annotation decision is labor-intensive and missing in most Natural Language Inference (NLI) datasets.
In this paper, we intend to examine whether Large Language Models can replace humans in generating these explanations. But how do we best query and select resulting model explanations? 
In this section, we detail the generation of model explanations, the selection of label-free and label-guided explanations, as illustrated in Figure~\ref{fig:autoex-generation}.

\paragraph{Model Explanation Generation}
We prompt LLMs to generate explanations for specific premise and hypothesis pairs and a given NLI label (see Table~\ref{tab:prompts-generation} in Appendix~\ref{app:prompts} for details).
Since NLI is subject to variation within the label \cite{DBLP:conf/emnlp/BowmanAPM15, DBLP:conf/emnlp/JiangTM23, DBLP:conf/acl/Weber-GenzelPMP24} and multiple annotators can agree on the same label for different reasons, we ask LLMs to list all possible explanations for each NLI label for a given instance.
We next introduce two explanation selection strategies: \textit{label-free} and \textit{label-guided}.

\paragraph{Label-Free Explanation Selection}
We first implement a baseline strategy, Label-Free Explanation Selection, by using one explanation for each of the three NLI labels: \texttt{\{Entailment, Neutral, Contradiction\}}.
This approach constructs three explanations across the three uniformly distributed NLI labels.
This label-free strategy benchmarks whether LLMs can approximate HJDs through diverse explanations without access to any label annotation. 

\paragraph{Label-Guided Explanation Selection}
A few NLI datasets have addressed the issue of human label variation (HLV) by including a small number of label annotations on the same instances, 5 for MNLI \cite{DBLP:conf/naacl/WilliamsNB18} and 4 for VariErr \cite{DBLP:conf/acl/Weber-GenzelPMP24}.  
Although earlier usages of these datasets are centered around the final label aggregated by majority voting, we consider using the small number of human labels as guidance to help build a combination of model explanations to approximate HJDs.
Unlike the label-free approach, we select explanations based on the annotated NLI labels for each instance. 
For example, for a given instance three explanations for \texttt{Entailment}, one for \texttt{Neutral}, and one for \texttt{Contradiction} are selected in Figure~\ref{fig:autoex-generation}.

\paragraph{Selecting First vs. Longest Explanations}
Since LLMs are prompted to exhaustively output explanations for given instances and labels, we propose two modes for selecting a desired number of explanations: one based on the linear order of LLM outputs (\textit{first}) and another based on the length in tokens of the output explanations (\textit{longest}).
For example, if two annotators have annotated an NLI instance with \texttt{Entailment}, we select the two longest model explanations that support \texttt{Entailment} under the \textit{longest} mode; or the initial two output \texttt{Entailment} explanations under the \textit{first} mode. 
The \textit{first} mode represents the primary preferences of LLMs, particularly in cases when prompting without explicit requirements to output all possible explanations.
The \textit{longest} mode can reveal more information regarding the reasoning between the premise and the hypothesis.\footnote{Note that the overall overlap rate between \textit{first} and \textit{longest} explanations are about 18.9\%. See Table~\ref{tab:stats_firstlong} in Appendix~\ref{app:exp1-detail-results} for detailed statistics.} Our experiments reveal that \textit{first} and \textit{longest} modes achieve similar results, cf.\ \S\ref{subsec:discussion-strategy}. 
Thus, in the main paper we report performances using \textit{longest} explanation(s); results with the \textit{first} mode are in the appendix.

\section{Can Model Explanations Help LLMs Approximate HJD as Humans Do?}\label{sec:llm-explanation-approximate-HJD}

Our first research question (\textbf{RQ1}) concerns \textit{whether LLM-generated explanations can model a human judgment distribution (HJD) as effectively as the human-written explanations}.

Previously, \citet{DBLP:conf/emnlp/Chen0PLKP24} introduced the task of approximating HJD from a few human-written labels and explanations using the LLM-based Model Judgment Distribution (MJD) Estimator.
We adopt their approach for our experiments and extend it from datasets that require human-written explanations to a broader range of datasets that include only human label annotations and benefit from LLM-generated explanations (\S\ref{sec:model-ex-generate}).

To validate the performance of \textbf{model-generated explanations} in approximating HJD, we prompt the MJD Estimator on gold NLI labels from the VariErr NLI dataset \citep{DBLP:conf/acl/Weber-GenzelPMP24} and LLM-generated explanations and first compare the results with those in \citet{DBLP:conf/emnlp/Chen0PLKP24} when both labels and explanations are human-written. 
We further experiment on the overlapping subset of a more widely-used dataset without explanations, MNLI \cite{DBLP:conf/naacl/WilliamsNB18} to show that our methodology generalizes well to more established datasets.
We present our experimental setups in \S\ref{subsec:exp1-setup}, and results in \S\ref{subsec:exp1-results}.

\subsection{Experimental Setup}
\label{subsec:exp1-setup}

\paragraph{MJD Estimator}

Following \citet{DBLP:conf/emnlp/Chen0PLKP24}, we estimate LLM's MJD through common-used multiple-choice question answering (MCQA) prompts~\cite{talmor-etal-2019-commonsenseqa,lin-etal-2022-truthfulqa,DBLP:conf/iclr/HendrycksBBZMSS21,DBLP:journals/tmlr/SrivastavaRRSAF23}.
We include in the prompts either (i) only the NLI instance, (ii) labels with human-written explanations, or (iii) labels with model-generated explanations. 
We then use the first-token probability method~\cite{DBLP:conf/icml/SanturkarDLLLH23,DBLP:journals/corr/abs-2306-16388,DBLP:journals/tmlr/LiangBLTSYZNWKN23} to obtain MJD.
To mitigate prompt bias (e.g. ``A preference''~\cite{DBLP:journals/corr/abs-2306-07951,DBLP:conf/iclr/Zheng0M0H24,DBLP:journals/tacl/TjuatjaCWTN24}, length bias and sequence bias), results reported in the main paper are averaged over ordering permutations of labels, explanations, and combinations. 
See Appendix~\ref{app:exp1-expset} for details.

\begin{table*}[t]
\centering
\resizebox{\textwidth}{!}{
\begin{tabular}{lccc|ccc|ccc|c}
\toprule
\multicolumn{1}{c}{\multirow{2}{*}{\textbf{Distributions}}} & \multicolumn{3}{c|}{\textbf{Dist. Comparison}} & \multicolumn{3}{c|}{\textbf{BERT Fine-Tuning Comparison (dev/test)}} & \multicolumn{3}{c|}{\textbf{RoBERTa Fine-Tuning Comparison (dev/test)}} &  \textbf{Global Metric}\\ \cmidrule(lr){2-11} 
\multicolumn{1}{c}{}          & {KL $\downarrow$} & {JSD $\downarrow$} & {TVD $\downarrow$}                                       & {KL $\downarrow$} &  {CE Loss $\downarrow$}  & {Weighted F1 $\uparrow$}    & {KL $\downarrow$}  &  {CE Loss $\downarrow$} & {Weighted F1 $\uparrow$} & {D.Corr $\uparrow$}    \\
\midrule
\multicolumn{11}{l}{\textit{Baseline from Human Annotations}}   \\
\midrule
 ChaosNLI HJD          & 0.000          & 0.000          & 0.000          & 0.073          / 0.077          & 0.967          / 0.974          & 0.645          / 0.609          & 0.062          / 0.060          & 0.933          / 0.922          & 0.696          / 0.653          & 1.000          \\
VariErr distribution       & 3.604          & 0.282          & 0.296          & 0.177          / 0.179          & 1.279          / 1.279          & 0.552          / 0.522          & 0.166          / 0.173          & 1.246          / 1.261          & 0.616          / 0.594          & 0.688          \\
MNLI distribution         & 1.242          & 0.281          & 0.295          & 0.104          / 0.100          & 1.062          / 1.042          & 0.569          / 0.555          & 0.101          / 0.093          & 1.052          / 1.020          & 0.625          / 0.607          & 0.795          \\
\midrule
\multicolumn{11}{l}{\textit{Model Judgment Distributions}}   \\
\midrule
Llama3       & 0.259          & 0.262          & 0.284          & 0.099          / 0.101          & 1.045          / 1.044          & 0.516          / 0.487          & 0.094          / 0.096          & 1.030          / 1.031          & 0.545          / 0.522          & 0.689          \\
\rowcolor{green!20}
+ human explanations & 0.238          & 0.250          & 0.269          & 0.098          / 0.099          & 1.043          / 1.039          & 0.575          / 0.556          & 0.091          / 0.092          & 1.021          / 1.019          & 0.641          / 0.616          & 0.771          \\
\multicolumn{11}{l}{{+ model explanations}} \\
\, \,  Label-Free               & 0.295          & 0.278          & 0.310          & 0.106          / 0.107          & 1.066          / 1.063          & 0.539          / 0.533          & 0.103          / 0.105          & 1.059          / 1.058          & 0.581          / 0.571          & 0.744          \\
\rowcolor{green!20}
\, \,  VariErr Label-Guided               & \textbf{0.234} & \textbf{0.247} & \textbf{0.266} & 0.097          / 0.098          & 1.041          / 1.037          & 0.558          / 0.544          & \textbf{0.089} / \textbf{0.091} & \textbf{1.016} / \textbf{1.014} & 0.633          / 0.626          & 0.760          \\
\, \,  MNLI Label-Guided              & 0.242          & 0.251          & 0.275          & \textbf{0.096} / \textbf{0.097}          & \textbf{1.037} / \textbf{1.034}          & \textbf{0.589}          / \textbf{0.580} & 0.090          / 0.092          & 1.019          / 1.018          & \textbf{0.657}          / \textbf{0.645}          & \textbf{0.849}
\\
\midrule
GPT-4o         & 0.265 & 0.263 & 0.289 & 0.103 / 0.096 & 1.059 / 1.029 & 0.526 / 0.517 & 0.093 / 0.092 & 1.027 / 1.018 & 0.525 / 0.521 & 0.703 \\
\rowcolor{green!20}
+ human explanations   & \textbf{0.187} & \textbf{0.207} & \textbf{0.223} & 0.093 / 0.098 & 1.027 / 1.036 & \textbf{0.570} / \textbf{0.552} & \textbf{0.079} / \textbf{0.080} & \textbf{0.986} / \textbf{0.987} & 0.617 / \textbf{0.617} & \textbf{0.769} \\
\multicolumn{11}{l}{{+ model explanations}} \\
\, \,  Label-Free            & 0.252 & 0.242 & 0.275 & 0.101 / 0.102 & 1.052 / 1.047 & 0.537 / 0.545 & 0.157 / 0.167 & 1.220 / 1.244 & 0.587 / 0.561 & 0.752 \\
\rowcolor{green!20}
\, \,  VariErr Label-Guided            & 0.192 & 0.209 & 0.226 & \textbf{0.092} / \textbf{0.093} & \textbf{1.026} / \textbf{1.022} & 0.554 / 0.551 & 0.088 / 0.089 & 1.013 / 1.008 & \textbf{0.618} / 0.598 & 0.761 \\
\bottomrule
\end{tabular}}
\caption{Evaluation results for measuring the closeness of MJD to HJD. The arrow points in the better direction. The bold numbers indicate the best results under the corresponding LLM. See Appendix~\ref{app:exp1-detail-results} for detailed results. }\label{tab:results-main}
\end{table*}

\paragraph{Datasets}

ChaosNLI~\cite{DBLP:conf/emnlp/NieZB20} includes 100 crowd-sourced annotations per instance and is considered gold Human Label Distribution (HJD) in our experiments. 
VariErr~\cite{DBLP:conf/acl/Weber-GenzelPMP24}, which tackles the explainability of NLI by asking a few experts to record the explanation behind each NLI label explicitly, includes 4 label and explanation annotations per instance on 341 items that overlap with ChaosNLI and MNLI \citep{DBLP:conf/naacl/WilliamsNB18}.
The overlapping subset of MNLI~\cite{DBLP:conf/naacl/WilliamsNB18} includes 5 label annotations per instance but without human-written explanations. 
Details are provided in The Appendix~\ref{app:exp1-expset} in Table~\ref{tab:nli_datasets}.

\paragraph{LLMs}

For both explanation generation and MJDs estimation, we utilized two open-source and one close-source instruction-tuned LLMs:
Llama3-Chat-70b~\cite{DBLP:journals/corr/abs-2407-21783}, 
Mixtral-8x7b-Instruct-v0.1~\cite{DBLP:journals/corr/abs-2401-04088}, 
and GPT-4o~\cite{DBLP:journals/corr/abs-2303-08774}.
We adopt the original chat templates for all models and set the parameter \texttt{do\_sample=False} in decoding (\texttt{temperature=0} for GPT-4o) to facilitate reproducibility.

\paragraph{Metrics}
Following \citet{DBLP:conf/emnlp/Chen0PLKP24}, we evaluated the MJDs on instance-level metrics,
Kullback-Leibler (KL) Divergence \cite{kullback1951information}, Jensen-Shannon Distance (JSD, \citealt{DBLP:journals/tit/EndresS03}) and Total Variation Distance (TVD, \citealt{DBLP:books/daglib/0018090}) as well as a global-level metric, 
Distance Correlation (D.Corr, \citealt{szekely2007measuring}).
We further used the MJDs as soft labels to fine-tune smaller language models using Cross-Entopy (CE) loss, namely, BERT \cite{DBLP:conf/naacl/DevlinCLT19} and RoBERTa \cite{DBLP:journals/corr/abs-1907-11692} and evaluated on remaining ChaosNLI dev/test sets using KL, Cross-Entropy Loss (CE Loss) and Weighted F1. 
The calculation formulas for all metrics can be found in the Appendix~\ref{app:exp1-expset}.

\subsection{Results}
\label{subsec:exp1-results}

\begin{figure*}[t]

        \centering
        \includegraphics[width=\textwidth]{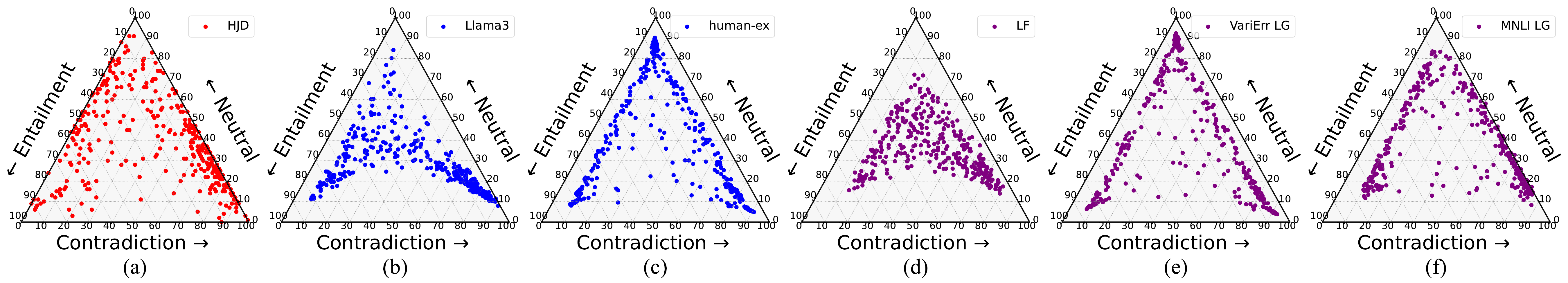}
        \caption{Llama3 Visualization using a ternary plot~\cite{gruber-etal-2024-labels}. Each point represents the label distribution for one NLI instance. From left to right: ChaosNLI HJD (red), Llama3 (blue), Llama3 with human explanations (blue), Llama3 with Label-Free (LF), Varierr/MNLI Label-Guided (LG) model explanations (3$\times$purple).}
        \label{fig:visual-main-Llama}
\end{figure*}

\begin{figure}[t]

        \centering
        \includegraphics[width=\linewidth]{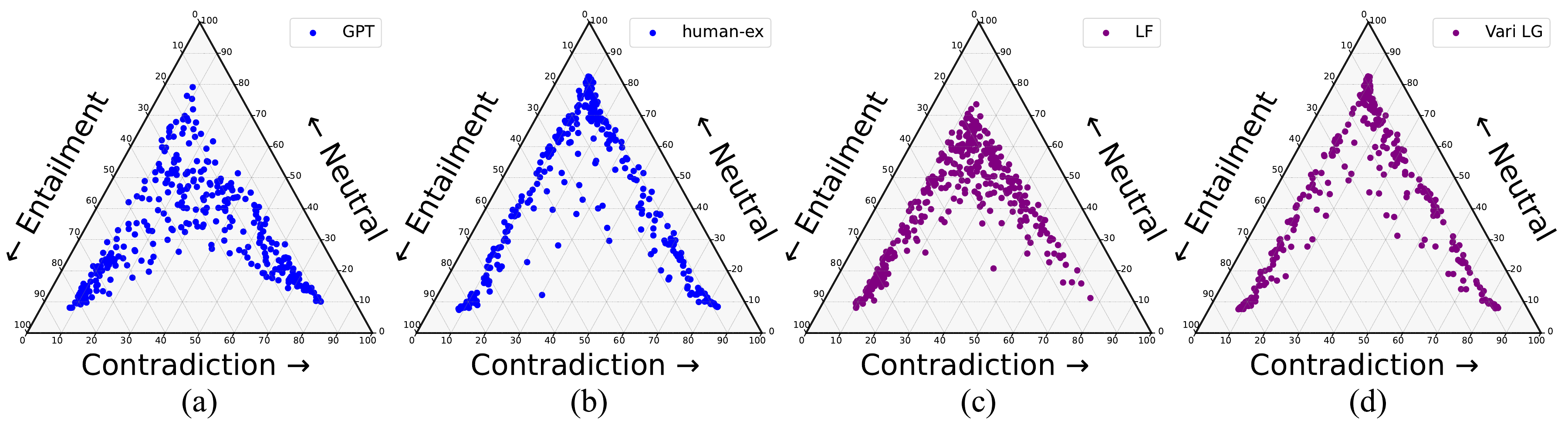}
        \caption{Visualization of MJDs from GPT-4o, with human explanations (2$\times$blue), with Label-Free and VariErr Label-Guided model explanations (2$\times$purple).}
        \label{fig:visual-main-GPT}
\end{figure}

Table~\ref{tab:results-main} presents our main results. 
The top panel shows the distribution comparison and fine-tuning outcomes on gold human label distributions.
We consider ChaosNLI HJD the ceiling, and the results of directly using VariErr and MNLI's multiple labels are baselines. 
For the MJD results in the bottom panel, see the three questions below. 
We also visualize the distributions of HJD and MJDs in Figures~\ref{fig:visual-main-Llama} and \ref{fig:visual-main-GPT} following~\citet{DBLP:conf/emnlp/Chen0PLKP24}.
Consistent with previous observations, Mixtral fails to capture HLV information from explanations. Further results and discussion on Mixtral are provided in the Appendix~\ref{app:exp1-detail-results-mixtral}.

\paragraph{Can LLMs approximate HJD via diverse model explanations without access to label annotation?}
For model explanations, we first explore whether eliminating human labeling and only entering one model explanation for each of the three possible classes is enough for LLMs to approximate the HJD.
The results show that Label-Free explanations perform poorly in distance comparisons when compared to LLMs, most likely due to the equal distribution of the three labels. 
However, the Label-Free method still scores significantly higher in the fine-tuning comparison and on D.Corr, and the latter two better reflect practical performance and global correlation.
As the Label-Free performance is between LLMs and LLMs with human explanations (including VariErr labels), results show that \textbf{model explanations do provide useful HLV information to allow LLMs to generate better MJDs}.
Figure~\ref{fig:visual-main-Llama}d plots the Label-Free (LF) MJD of Llama3. 
Its distribution is relatively smooth, uniform, and less centered on particular labels than the original MJD of Llama3 in Figure~\ref{fig:visual-main-Llama}b.
We also visualized the MJDs from GPT-4o.
Unlike Llama3, the original MJD of GPT-4o in Figure~\ref{fig:visual-main-GPT}a is relatively uniform, and Label-Free model explanations help move some wrong points upward (in NLI, more disagreements are found between \texttt{E-N} and \texttt{N-C}, and fewer are found between \texttt{E-C}, the bottom points in the triangle). This may explain why Label-Free can be better than LLMs on the fine-tuning comparison and D.Corr.

\paragraph{Are model explanations comparable to human's when helping LLMs to approximate HJD?}
We take a step back and consider using a few gold human labels that are easy to obtain for most datasets but pair them with LLM-generated explanations.
The green rows in Table \ref{tab:results-main} show these comparative settings, where the only difference is whether the explanations are human-annotated or LLM-generated. The results show that \textbf{the VariErr Label-Guided model explanation achieves comparable results to LLMs with human explanations}, using the same human labels from VariErr NLI. 
The visualizations in Figure~\ref{fig:visual-main-Llama}e and Figure~\ref{fig:visual-main-GPT}d illustrate that the MJDs of VariErr-guided shows a distribution similar to human explanations (Figure~\ref{fig:visual-main-Llama}c of Llama3 and Figure~\ref{fig:visual-main-GPT}b of GPT). More detailed comparative analysis and ablation experiments are provided in Section~\ref{sec:exp3}.

\paragraph{Does our approach extend to NLI datasets that do not have human-provided explanations?}
To investigate the generalizability of our approach,\footnote{Considering the experimental cost, we only tested this setting on open-sourced LLMs. Mixtral results in Appendix~\ref{app:exp1-detail-results-mixtral}.}
we used MNLI-guided explanation generation utilizing the 5 NLI labels for each MNLI instance in the overlapping NLI data subset.\footnote{A preliminary experiment on 59 NLI instances with 5 explanation-label pairs from VariErr found that varying number of explanations between 3 to 5 does not have much impact.}
Table~\ref{tab:results-main} shows that
MNLI-guided achieves the best MJD on F1 for both BERT and RoBERTa FT, as well as on D.Corr.
Figure~\ref{fig:visual-main-Llama}f visualizes the MJD of MNLI-guided, and its distribution is more similar to Chaos NLI HJD. Both are smoother at the upper corner of the triangle than human (Figure~\ref{fig:visual-main-Llama}c) or VariErr-guided explanations (Figure~\ref{fig:visual-main-Llama}e), and are also more skewed towards the contradiction side than Label-Free (Figure~\ref{fig:visual-main-Llama}d) and Llama3 itself (Figure~\ref{fig:visual-main-Llama}b).
The comparable performance of VariErr and MNLI-guided explanations to human explanations shows \textbf{the scalability of our model-generated explanations}.

\section{Can Model-Generated Explanations Enhance Performance on OOD Data?}
\label{sec:exp2}

The overlapping instances in ChaosNLI, VariErr, and MNLI allow us to compare human-written and model-generated explanations in HJD estimation and fine-tuning.
Our second research question (\textbf{RQ2}) is \textit{whether the generated MJDs can help downstream language models solve other difficult NLI tasks out-of-domain (OOD)}.

\subsection{Experiment Setup}\label{subsec:exp2-setup}

\paragraph{Dataset}
The ANLI dataset ~\cite{DBLP:conf/acl/NieWDBWK20} is a challenging NLI dataset collected by an adversarial procedure.
Mechanical Turkers are instructed to continue writing hypotheses for a given context and target label until a trained BERT/RoBERTa model (using MNLI, SNLI, etc.) outputs a wrong label prediction.
This iterative process is conducted on three rounds (R1-R3) of annotations, and each round contains different context texts, mainly from Wikipedia, but R3 includes additional news, fiction, speech, and other contexts. 
We conduct OOD evaluation on R1-R3 data of the ANLI test set.\!\footnote{We discuss the \emph{OOD} definition and data contamination risk of ANLI in Appendix~\ref{app:data-contamination}.} 

\paragraph{Models}
Since ANLI is OOD and gold HJD is inaccessible, we leverage all the fine-tuned BERT and RoBERTa models from \S\ref{sec:llm-explanation-approximate-HJD} as classifiers and directly evaluate them on the ANLI test set.

\begin{table}[t]
\centering
\resizebox{\linewidth}{!}{
\begin{tabular}{lccc|ccc}
\toprule
\multicolumn{1}{c}{\multirow{2}{*}{\textbf{Classifiers}}} & \multicolumn{3}{c|}{\textbf{BERT FT Test}} & \multicolumn{3}{c}{\textbf{RoBERTa FT Test}} \\ \cmidrule(lr){2-7} 
\multicolumn{1}{c}{}              & {R1 $\uparrow$} &  {R2 $\uparrow$}  & {R3 $\uparrow$}   & {R1 $\uparrow$} &  {R2 $\uparrow$}  & {R3 $\uparrow$}   \\
\midrule
\multicolumn{7}{l}{\textit{Classifiers without distribution training}}   \\
\midrule
Out-of-the-box LM        & 0.170          & 0.176          & 0.197          & 0.167          & 0.167          & 0.168          \\
MNLI-FT-LM         & 0.220          & 0.269          & 0.293          & 0.292          & 0.262          & 0.257          \\
\midrule
\multicolumn{7}{l}{\textit{Classifiers trained on label distributions
}}   \\
\midrule
ChaosNLI HJD           & 0.268          & 0.289          & 0.332          & 0.357          & 0.331 & 0.338          \\
\rowcolor{green!20}
VariErr distribution      & 0.302          & 0.259          & 0.319          & 0.402          & 0.311          & 0.321          \\
\rowcolor{yellow!20}
MNLI distribution          & 0.229          & 0.260          & 0.279          & 0.317          & 0.275          & 0.281          \\
\midrule
\multicolumn{7}{l}{\textit{Classifiers trained on MJDs
}}   \\
\midrule
Llama3       & 0.246          & 0.276          & 0.306          & 0.304          & 0.297          & 0.304          \\
\rowcolor{green!20}
+ human explanations & 0.296          & 0.289          & 0.349          & 0.400          & \textbf{0.330}          & \textbf{0.344} \\
\multicolumn{7}{l}{{+ model explanations}}   \\
\, \, Label-Free               & 0.292          & \textbf{0.295} & 0.328          & 0.314          & 0.262          & 0.323          \\
\rowcolor{green!20}
\, \, VariErr Label-Guided               & \textbf{0.305}          & 0.285          & \textbf{0.349} & \textbf{0.411} & 0.324          & 0.319          \\
\rowcolor{yellow!20}
\, \, MNLI Label-Guided              & 0.284          & 0.283          & 0.321          & 0.339          & 0.287          & 0.307      
\\
\midrule
GPT-4o       & 0.258 & 0.263 & 0.295 & 0.309 & 0.282 & 0.302 \\
\rowcolor{green!20}
+ human explanations     & \textbf{0.351} & \textbf{0.294} & \textbf{0.332} & \textbf{0.393} & \textbf{0.324} & \textbf{0.325} \\
\multicolumn{7}{l}{{+ model explanations}}   \\
\, \, Label-Free   & 0.285 & 0.283 & 0.315 & 0.350 & 0.282 & 0.310 \\
\rowcolor{green!20}
\, \, VariErr Label-Guided    & 0.341 & 0.293 & 0.330 & 0.393 & 0.324 & 0.323 \\
\bottomrule
\end{tabular}}
\caption{ANLI test results as Weighted F1 scores. Bold numbers indicate the best results under the corresponding LLM. Detailed results for individual runs as well as Mixtral's performance are elaborated in Appendix~\ref{app:anli}.}\label{tab:results-main-ANLI}
\end{table}

\subsection{Results}\label{subsec:exp2-result}
Results are shown in Table~\ref{tab:results-main-ANLI}.
The out-of-the-box BERT and RoBERTa models perform badly on ANLI.
After fine-tuning on the MNLI training set via majority label classification, the classifiers improved slightly, similar to results reported in \citet{DBLP:conf/acl/NieWDBWK20} trained on both SNLI and MNLI. 

We also evaluate classifiers fine-tuned on ChaosNLI, VariErr, and MNLI human label distributions and found that all scores improved compared to earlier distribution-less training.
This further substantiates the significance of human label distribution in enhancing the robustness and generalization of the model. 
It is worth noting that classifiers trained with VariErr distribution perform better than classifiers trained with MNLI distribution on all test sets, and even outperform the ChaosNLI HJD in the RoBERTa R1 setting. 
One hypothesis could be that the ChaosNLI HJD with 100 annotations, though more informative on the 341 instances, is out-of-distribution and thus less suitable for modeling label distributions in ANLI. 
Furthermore, we fine-tuned the classifiers with MJDs from LLMs.
For Llama3, all MJDs with the help of explanations, whether from humans or models, have improved performance compared to MJDs trained without explanations, consistent with Table~\ref{tab:results-main}.

If we look at the label distribution sources, the results can be divided into two categories: green from VariErr and yellow from MNLI. 
With the help of explanations, the results of MJDs all exceed the corresponding label distributions.
Moreover, the green rows consistently perform better than the yellow rows for both label distribution and  MJDs trained with explanations.
We hypothesize that the quality of the datasets matters since VariErr is collected from expert linguists whereas MNLI from crowd workers. 
Overall, results show that \textbf{MJDs generated by our method are robust on OOD datasets without label distributions or explanations}.

\section{Human versus Model Explanations: Are They Different and Does It Matter?}
\label{sec:exp3}

\begin{figure*}[t]

        \centering
        \includegraphics[width=\textwidth]{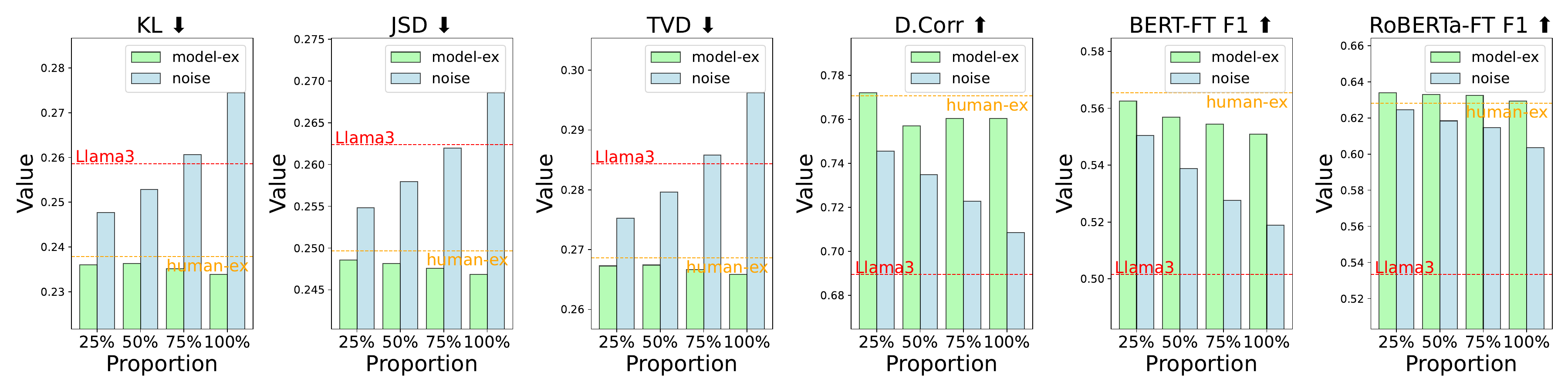}
        \caption{Results for ablation study. The green bars represent the performance of MJDs when replaced by model-generated explanations on the same instance and label, while the blue bars represent that with noise replacements. The orange/red dashed line shows the performance of Llama3 with/without full human explanations.}
        \label{fig:ablation-bar-replace-long}
\end{figure*}

\begin{figure}[htbp]
	\centering
	\begin{subfigure}{\linewidth}
		\centering
		\includegraphics[width=\linewidth]{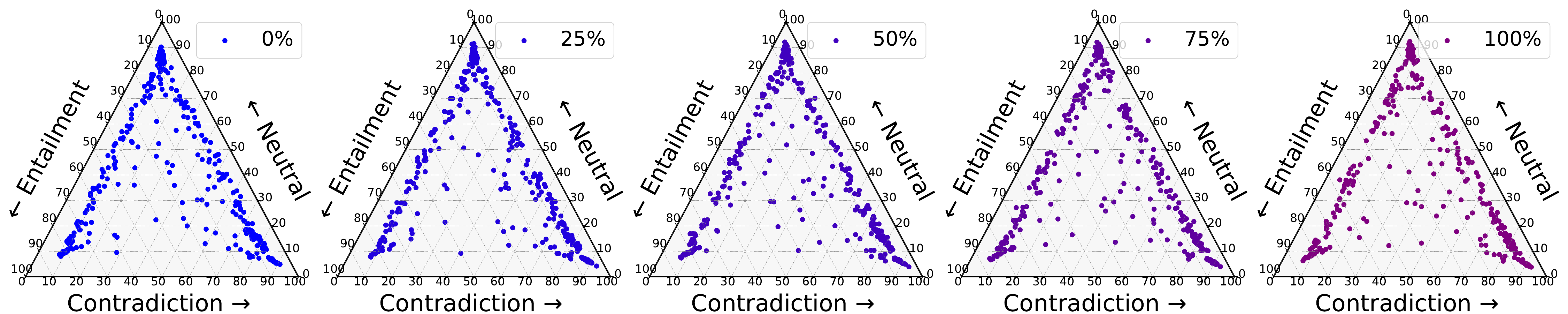}
		\caption{Gradually replaced by model explanations.}
		\label{fig:ablation-visual-long-auto}
	\end{subfigure}

	\centering
	\begin{subfigure}{\linewidth}
		\centering
		\includegraphics[width=\linewidth]{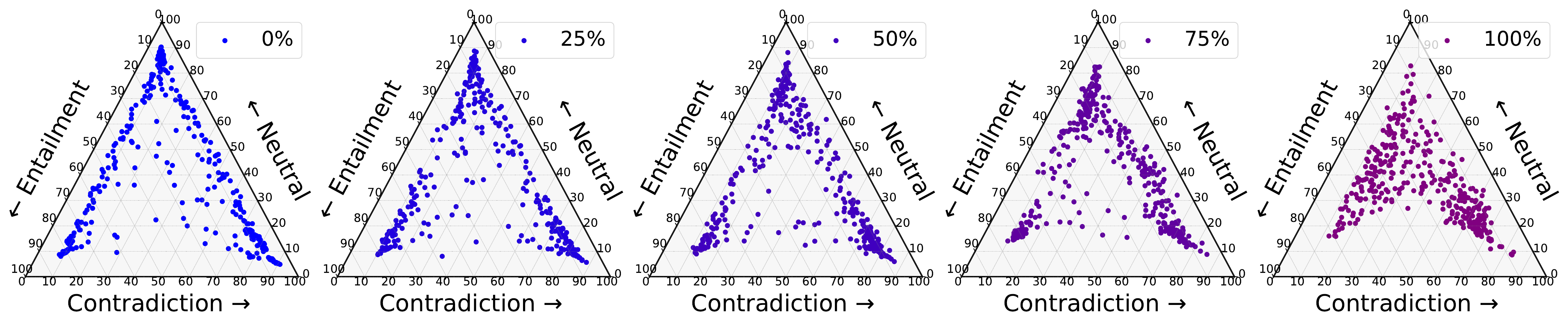}
		\caption{Gradually replaced by noise explanations.}
		\label{fig:ablation-visual-long-noise}
	\end{subfigure}
	\caption{Visualization of gradually replacing human explanations with model or noise explanations.}
	\label{fig:ablation-visual}
\end{figure}

In addition to the consistent performance gain of using model or human explanations for HJD estimation, we are interested in delving into the nuances between model and human explanations.
We decompose this goal into two questions (\textbf{RQ3\&4}):

\begin{itemize}
\item  \textit{How does the performance of MJDs change as we gradually replace human explanations with model explanations?}
\item \textit{Does the content of model explanations matter, or do the human labels play a decisive role?}
\end{itemize}

Since Llama3 is consistently better compared to Mixtral and GPT4,\footnote{Similarly, \citet{DBLP:conf/emnlp/Chen0PLKP24} found that Llama3 outperformed Mixtral on approximating HJD.} we conduct the following ablation studies only on Llama3.
We start from MJDs generated by all human explanations (0\% replacement) and gradually increase the proportion of model explanations by replacing human explanations with model ones, until all explanations are model-generated (100\% replacement). 
We then evaluate the performance on different explanation replacement rates.
To explore whether the content of the model explanations matters, we design a controlled \textit{noise replacement} experiment where we replace a human-written explanation with an irrelevant model explanation from another NLI instance but with support for the same \texttt{Entailment/Neutral/Contradiction} label.

Figure~\ref{fig:ablation-bar-replace-long} plots the gradual replacement of VariErr's four human explanations by model or noise explanations on six metrics (scores in Appendix~\ref{app:ablation}).
The key finding is that \textbf{model and human explanations result in similar performance, while noise replacement clearly hurts}. In more detail, we can observe that 
when gradually replacing human explanations with model ones, fluctuations are small on all metrics compared to full human explanations.
Importantly, noise replacements deteriorate performances significantly, resulting in remarkably lower F1 scores and higher distribution divergence. These results provide further evidence that generated explanations are a viable alternative to more costly human-written explanations. 

Additionally, we see in Figure~\ref{fig:ablation-visual-long-auto} that the shape of MJDs remains similar when gradually replaced by model explanations. 
In contrast, in Figure~\ref{fig:ablation-visual-long-noise} when replaced by noise, the distribution gradually loses its shape and gathers in the center.
The bar and ternary plots above agree that \textbf{the relevant contents of human or model explanations are crucial} in addition to the guidance of human labels.

\begin{table}[t]
\centering
\resizebox{\linewidth}{!}{
\begin{tabular}{lccc|c}
\toprule
\multicolumn{1}{c}{\multirow{2}{*}{\textbf{Distributions}}} & \multicolumn{3}{c|}{\textbf{Dist. Comparison}} &  \textbf{Global Metric}\\ \cmidrule(lr){2-5} 
\multicolumn{1}{c}{}          & {KL $\downarrow$} & {JSD $\downarrow$} & {TVD $\downarrow$}                                      & {D.Corr $\uparrow$}    \\
\midrule
VariErr distribution       & 6.628          & 0.357          & 0.352          & 0.907          \\
Llama3 MJD       & 0.029          & 0.068          & 0.088          & 0.691          \\
\rowcolor{green!20}
+ human explanations & 0.000          & 0.000          & 0.000          & 1.000          \\
\multicolumn{5}{l}{{+ replace model explanations}} \\
\, \, Label-Free 100\%              & 0.024          & 0.067          & 0.088          & 0.647          \\
\, \, VariErr Label-Guided 25\%     & \textbf{0.001} & \textbf{0.012} & \textbf{0.015} & \textbf{0.977} \\
\, \, VariErr Label-Guided 50\%     & 0.003          & 0.017          & 0.022          & 0.959          \\
\, \, VariErr Label-Guided 75\%     & 0.003          & 0.019          & 0.024          & 0.950          \\
\, \, VariErr Label-Guided 100\%     & 0.004          & 0.021          & 0.027          & 0.939        
\\
\bottomrule
\end{tabular}}
\caption{Results from a human-explanation-centric view. All MJDs are compared to the MJD in the green row. }\label{tab:results-ablation-centric}
\end{table}

\paragraph{Switching to a human-explanation-centric view}
All comparisons above treat the ChaosNLI HJD as the comparison target to explore \textbf{RQ3}. 
It would be useful to temporarily alter our perspective and treat the MJD with human explanations as the target, highlighted in green in Table~\ref{tab:results-ablation-centric}. 
We observe that on all metrics, Llama3 without explanation and with Label-Free are far from MJD with human explanations. 
Gradually replacing by model explanations keeps the generated MJDs slightly away from the centric but remains very similar.\footnote{We examine the similarities between model/human explanations following~\citet{giulianelli-etal-2023-comes} (Table~\ref{tab:results-ablation-similarity-parallel} in Appendix~\ref{app:ablation}). Our observation mirrors theirs in that model explanations differ moderately from human explanations regarding lexicon, syntax, and semantics.
Nevertheless, LLMs still found a way to obtain comparable information for modeling HJD.}
It is worth noting that although VariErr's label distribution is far considering traditional instance-level distance comparison metrics, it is relatively similar in the global metric D.Corr. 
This observation is consistent with that VariErr distribution performs well in FT comparison in Table~\ref{tab:results-main}, further corroborating
the finding by \citet{DBLP:conf/emnlp/Chen0PLKP24} on the suitability of D.Corr in comparing label distributions.

\begin{table*}[t]
\centering
\resizebox{\textwidth}{!}{
\begin{tabular}{lccc|ccc|ccc|c}
\toprule
\multicolumn{1}{c}{\multirow{2}{*}{\textbf{Distributions}}} & \multicolumn{3}{c|}{\textbf{Dist. Comparison}} & \multicolumn{3}{c|}{\textbf{BERT Fine-Tuning Comparison(dev/test)}} & \multicolumn{3}{c|}{\textbf{RoBERTa Fine-Tuning Comparison(dev/test)}} &  \textbf{Global Metric}\\ \cmidrule(lr){2-11} 
\multicolumn{1}{c}{}          & {KL $\downarrow$} & {JSD $\downarrow$} & {TVD $\downarrow$}                                       & {KL $\downarrow$} &  {CE Loss $\downarrow$}  & {Weighted F1 $\uparrow$}    & {KL $\downarrow$}  &  {CE Loss $\downarrow$} & {Weighted F1 $\uparrow$} & {D.Corr $\uparrow$}    \\
\midrule
Llama3          & 0.258          & 0.261          & 0.286          & 0.092          / 0.093          & 1.024          / 1.020          & 0.514          / 0.471          & 0.092          / 0.095          & 1.025          / 1.026          & 0.531          / 0.512          & 0.684          \\
+ human explanations    & 0.240          & 0.249          & 0.275          & 0.090          / 0.090          & 1.017          / 1.011          & 0.594          / 0.567          & 0.089          / 0.091          & 1.014          / 1.015          & 0.618          / 0.597          & 0.750          \\
\multicolumn{11}{l}{{+ replace \textit{preferred} model explanations}}   \\
\, greedy 75.75\%        & 0.241          & 0.248          & 0.274          & 0.089          / 0.090          & 1.017          / 1.011          & 0.584          / 0.569          & 0.088          / 0.090          & 1.013          / 1.013          & 0.619          / 0.594          & 0.733          \\
\, representative 55.25\%      & 0.240          & 0.248          & 0.274          & 0.089          / 0.090          & 1.016          / 1.011          & 0.587          / 0.567          & 0.088          / 0.091          & 1.013          / 1.014          & 0.619          / 0.597          & 0.739          \\
\multicolumn{11}{l}{{+ replace \textit{unpreferred} model explanations}}   \\
\, greedy 68.5\%         & 0.239          & 0.247          & 0.273          & \textbf{0.089}          / \textbf{0.089} & \textbf{1.016}          / \textbf{1.009} & \textbf{0.589} / \textbf{0.571} & \textbf{0.087} / 0.090          & \textbf{1.011} / 1.012          & \textbf{0.623} / 0.599          & 0.752          \\
\, representative 63.25\%      & \textbf{0.237} & \textbf{0.246}          & \textbf{0.271}          & 0.089          / 0.089          & 1.016          / 1.010          & 0.584          / 0.566          & 0.088          / \textbf{0.090} & 1.011          / \textbf{1.012} & 0.621          / \textbf{0.607} & \textbf{0.761} 

\\
\bottomrule
\end{tabular}}
\caption{Contrastive results using annotator-preferred vs. unpreferred explanations on 100 validated NLI instances. \textit{greedy} substitutes human explanations with as many model explanations as possible, while \textit{representative} substitutes per NLI label: for each attested \texttt{E/N/C} label, replace by at most one model explanation (if any). }\label{tab:results-validation}
\end{table*}

\section{Can Human Preference Lead to Better Explanation Selection?}
\label{subsec:discussion-strategy}

As illustrated in \S\ref{sec:model-ex-generate}, we adopted two intuitive strategies to select model explanations:
\textit{first} and \textit{longest}. 
Across all experiments there is no significant difference in the results obtained from the two modes (\textit{first} results in Appendix).
To manually assess the quality of these model explanations,  we recruited a human annotator to validate 1,581 Llama-generated explanations on 100 NLI instances. 
Two questions were asked for each model explanation: 
(1) \textit{Does the model explanation faithfully describe the meanings of the premise and hypothesis (yes/no)?}  
and (2) \textit{Does the explanation bring additional relevant information to support a reasonable NLI Label (yes/no)? If yes, what is the label (E/N/C)?}
The first question allows the annotator to filter out model explanations including factual errors or hallucinations. 
The second question asks for the relevance and logical reasoning of the model explanation, but depending on individuals' world knowledge, it may reflect the annotator's preference. 
When an annotator classifies a model explanation as reasonable, we regard it as a \textit{preferred} explanation and replace its NLI label with the annotator's label (if different).
When a model explanation is 
classified as \textit{unpreferred}, we keep the original NLI label. 

\begin{table}[t]
\centering
\resizebox{\linewidth}{!}{
\begin{tabular}{lccc|ccc|cc|c}
\toprule
\multicolumn{1}{c}{\multirow{2}{*}{\textbf{Datasets}}} & \multicolumn{3}{c|}{\textbf{Lexical}} & \multicolumn{3}{c|}{\textbf{Syntactic}} & \multicolumn{2}{c|}{\textbf{Semantic}} &  \textbf{AVG}\\ \cmidrule(lr){2-10} 
\multicolumn{1}{c}{}          & {n = 1$\downarrow$} & {n = 2 $\downarrow$} & {n = 3$\downarrow$}                                     & {n = 1$\downarrow$ } & {n = 2$\downarrow$ } & {n = 3$\downarrow$ }         & {Cos.$\downarrow$ }  &  {Euc.$\downarrow$ }  & {AVG $\downarrow$}    \\
\midrule
human-ex         & 0.335 & 0.098 & 0.042 & 0.767 & 0.341 & 0.140 & 0.528 & 0.520 & 0.428 \\
\midrule
\multicolumn{10}{l}{{replaced \textit{preferred} model explanations}}   \\
\, greedy   & 0.416 & 0.157 & 0.082 & 0.874 & 0.488 & 0.233 & 0.540 & 0.532 & 0.474 \\
\, represent. & 0.392 & 0.149 & 0.089 & 0.835 & 0.426 & 0.205 & 0.542 & 0.541 & 0.466 \\
\midrule
\multicolumn{10}{l}{{replaced \textit{unpreferred} model explanations}}   \\  
\, greedy    & 0.387 & 0.130 & 0.069 & 0.841 & 0.432 & 0.196 & 0.527 & 0.528 & 0.457 \\
\, represent.  & 0.378 & 0.130 & 0.073 & 0.837 & 0.426 & 0.195 & 0.534 & 0.532 & \textbf{0.455}
\\
\bottomrule
\end{tabular}}
\caption{Results for linguistic variability check. }\label{tab:results-6-similarity-centric-validation}
\end{table}

Results are in Table~\ref{tab:results-validation} (details in Appendix~\ref{app:validation}). 
Surprisingly, model explanations from the \textit{unpreferred} set achieved the best results on most metrics, which is different from our original expectation.
This may be due to the \textit{unpreferred} explanations being more diverse than the \textit{preferred} ones (note that this is limited to the judgment of a single annotator). 
To verify this hypothesis, we average similarities among each pair of explanations ($C\binom {2}{n}$ pairs for $n$ explanations) on each NLI instance.
Table~\ref{tab:results-6-similarity-centric-validation} shows that human explanations have the greatest variability on all using \citet{giulianelli-etal-2023-comes}'s similarity measures (the lower the value, the higher the variability).
Moreover, model explanations that counter the preferences of one human annotator (\textit{unpreferred}) have a higher variability, providing more diverse perspectives, which aligns with the observation in Table~\ref{tab:results-validation}.
These experiments show the potential of \textit{variability} as a metric for measuring the model explanations when helping LLMs approximate HJD. 
We expanded this variability check to the main experiment.
Results in Table~\ref{tab:results-similarity-centric-main} in Appendix~\ref{app:validation} show that MNLI-guided explanations from Llama3 have the best variability, which is consistent with the results in Table~\ref{tab:results-main}.
However, variability cannot be directly linked to the main results in all circumstances, cf.\ Appendix~\ref{app:validation} for a discussion.

\section{Related Work}

\paragraph{LLMs to generate explanations} 
Recently LLM-generated explanations have been used in various tasks~\cite[e.g., reasoning, sentimental analysis, recommender systems, education, abusive language detection;][]{DBLP:journals/corr/abs-2210-06726,DBLP:journals/corr/abs-2310-11207,DBLP:conf/um/LubosTFEL24,DBLP:conf/educon/Abu-Rasheed0F24,0bf90dafe25b49628b7328f6834c2ee1}.
\citet{DBLP:journals/corr/abs-2402-10532} find that LLM-generated explanations show selectivity and contain illustrative elements, but less frequently are subjective or misleading.
\citet{DBLP:conf/chi/Wang0RMM24} utilize human feedback on LLM-generated explanations to assist in collaborative annotation. This is even a potential future direction for us—to help build a human-LLM collaboration explanation NLI dataset. \citet{DBLP:conf/naacl/WiegreffeHSRC22} and \citet{DBLP:conf/acl/MishraRM0H24} focus on evaluating the reasonableness, conciseness, and other aspects of LLM-generated explanations from a human-label-guided assessment perspective. 
Unlike previous methods, we guide LLMs to generate more diverse explanations for NLI to analyze HLV.

\paragraph{LLMs to model label distributions}
Despite the increasing promise of LLMs as annotators, many studies have attempted to use LLMs to approximate label distributions, with mixed success~\cite{DBLP:journals/corr/abs-2305-14770,pavlovic-poesio-2024-effectiveness,DBLP:conf/emnlp/LeeAT23,madaan2024lost}.
\citet{DBLP:conf/emnlp/Chen0PLKP24} combine human explanations and labels to enhance LLMs' performance to approximate HJD, but rely on datasets with human-provided explanations. In contrast, we are the first to leverage LLM-generated explanations to model HJD, addressing the scarcity of explanation datasets.

\section{Conclusion}

This paper demonstrates that large language models can effectively generate explanations to approximate human judgment distribution in NLI. Our experiments reveal that model-generated explanations, when combined with a few human labels, yield results comparable to human-provided explanations in approximating HJD. Notably, our approach generalizes to explanation-free datasets and remains effective in challenging OOD test sets. 
Results indicate that LLM-generated explanations can significantly reduce annotation costs, making it a scalable and efficient proxy for capturing HLV.

\section*{Limitations}

One limitation of this work is that our current method only considers explanations generated by one LLM to help itself achieve better MJD. Exploring a cross-LLM approach that combines explanations from multiple LLMs could be interesting, as it may provide more diverse perspectives. However, we leave this investigation for future work, as the scope of this paper is to demonstrate that LLM-generated explanations are as effective as human explanations in helping LLMs approximate HJD.

Another area for improvement lies in the method used to obtain the LLM’s opinion distribution. Currently, we rely on the MCQA prompt combined with the first-token-probability method to derive MJD, applying basic normalization or softmax as the transformation function to convert logits into probabilities. This method may not be universally suitable for all LLMs. Exploring alternative approaches to better capture distributed opinions from LLMs is an intriguing direction for future work.

Besides, we also discuss the limitation of the explanation selection strategy presented in \S\ref{sec:model-ex-generate}. The "first" explanation was chosen because it could reflect the initial intuition or immediate reasoning that a model or human might provide, offering a baseline for interpretability. The "longest" explanation was included to capture more detailed or comprehensive reasoning, which may provide deeper insights into the model's decision-making process. While these characteristics serve as a useful starting point for analysis, we acknowledge that they represent only a subset of the dimensions that could be explored when evaluating explanations.
Thus, how explanation diversity and quality influence the performance of HJD approximation is still up to future research.

\section*{Acknowledgements}
We thank the members of the MaiNLP lab for their insightful feedback on earlier drafts of this paper. 
We specifically appreciate the suggestions of Xinpeng Wang, Robert Litschko, Philipp Mondorf, Bolei Ma, Soh-Eun Shim, Felicia Körner, Andreas Säuberli, Verena Blaschke, Shijia Zhou, Diego Frassinelli, Yang Janet Liu, Michael A. Hedderich, and Rob van der Goot. 
We extend our gratitude to Huangyan Shan for annotating the human preference experiment.
We are also grateful to the anonymous reviewers for their constructive feedback. 
BC acknowledges his membership in the European Laboratory for Learning and Intelligent Systems (ELLIS) PhD program.
This research is supported by ERC Consolidator Grant DIALECT 101043235
and the UK Research and Innovation (UKRI) Frontier Research Grant EP/Y031350/1 EQUATE (the UK government's funding guarantee for ERC Advanced Grants).


\appendix

\section{Prompt for Generating Model Explanations}
\label{app:prompts}

Table~\ref{tab:prompts-generation} illustrates the prompt we used to let LLMs generate model explanations. For the LLM with the ``system'' role in the chat template, we choose not to set the content of the ``system'' role to be consistent with other LLMs.

\begin{table}[t]
\scriptsize
\centering
\begin{tabular}{P{0.09\textwidth} | P{0.33\textwidth}}
\toprule 
\multicolumn{1}{l|}{\textbf{Function}} & \multicolumn{1}{l}{\textbf{General Instruction Prompt}} \\
\midrule
model explanation generation & \textbf{"role": "user", "content"}:  \newline  You are an expert in Natural Language Inference (NLI). Please list all possible explanations for why the following statement is \{relationship\} given the context below without introductory phrases. \newline Context: \{premise\} \newline Statement: \{hypothesis\} \newline Answer:                                 \\
\bottomrule
\end{tabular}

\caption{Instruction prompt for LLMs to generate model explanations. \texttt{relationship} is one of \{\texttt{true (entailment)}, \texttt{undetermined (neutral)}, \texttt{false (contradiction)}\}.
}
\label{tab:prompts-generation}
\end{table}
\section{Details for the Main Experiment}
\label{app:main-results}

In this Section, we first illustrate the detailed experimental settings in \S\ref{app:exp1-expset} for \S\ref{subsec:exp1-setup}, and then we report detailed scores in \S\ref{app:exp1-detail-results} for \S\ref{subsec:exp1-results}. We also include Mixtral's results in \S\ref{app:exp1-detail-results-mixtral}.

\subsection{Experimental Settings Details}
\label{app:exp1-expset}

\begin{table}[t]
\scriptsize
\centering
\begin{tabular}{P{0.09\textwidth} | P{0.33\textwidth}}
\toprule 
\multicolumn{1}{l|}{\textbf{Function}} & \multicolumn{1}{l}{\textbf{General Instruction Prompt}} \\
\midrule
LLM original & \textbf{"role": "user", "content"}:  \newline  Please determine whether the following Statement is true (entailment), undetermined (neutral), or false (contradiction) given the Context below and select ONE of the listed options and start your answer with a single letter. \newline Context: \{premise\} \newline Statement: \{hypothesis\} \newline A. Entailment \newline B. Neutral \newline C. Contradiction. \newline Answer:                                 \\
\midrule  
LLM with explanations & \textbf{"role": "user", "content"}:  \newline  Please carefully and fairly base your selection on the comments below to determine whether the following Statement is true (entailment), undetermined (neutral), or false (contradiction) given the Context below and select ONE of the listed options and start your answer with a single letter. \newline Context: \{premise\} \newline Statement: \{hypothesis\} \newline Comment 1: \{explanation 1\}, so I choose \{label 1\} \newline Comment 2: \{explanation 2\}, so I choose \{label 2\} \newline ... \newline A. Entailment \newline B. Neutral \newline C. Contradiction. \newline Answer:                                 \\
\bottomrule
\end{tabular}

\caption{Instruction prompt of different types to transform NLI into a multi-choice question format.}
\label{tab:prompts-MJD}
\end{table}

\paragraph{MJD Estimator}
We adopt the MJD Estimator following~\citet{DBLP:conf/emnlp/Chen0PLKP24} to generate model judgment distributions from LLMs. Model/human explanations are combined with labels together with NLI instance to fill in the MCQA prompt as shown in Table~\ref{tab:prompts-MJD}.
To capture the original opinion from the LLM, \texttt{LLM original} directly inputs the content of the NLI instance and asks the LLM for its choice; to capture the LLM's perspective influenced by human annotations, \texttt{LLM with explanations} incorporates human explanations of label choices as ``comments'', which are placed after the NLI instance but before the MCQA part.

With the input prompt above, we next map LLMs’ output from \texttt{[A,B,C]} to probabilities as model judgment distributions. We leverage the logits of the first output token before the decoding process, and extract the three scores corresponding to \texttt{[A,B,C]}.
Via normalization or softmax function, we can transform these scores into probabilities, which is considered as model judgment distribution that represents the label distributions among \texttt{[Entailment, Neutral, Contradiction]}.

For Llama3 we adopt the normalization method as all output logits are positive, thus we can avoid the influence of parameters as much as possible, because the normalization transformation does not introduce additional parameters.
However, as negative logits exist from GPT-4o and Mixtral, we have to leverage softmax transformation to get the label distributions. 
Some recent work has discussed the impact of temperature \(\tau\) in the softmax function on HLV observation~\cite{pavlovic2024understanding}, but we try not to discuss this variable too much because it is not the focus of this paper. Without loss of generality, we perform a \(\tau = 10\) softmax transformation on all GPT-4o logits and a \(\tau = 20\) softmax transformation on all Mixtral logits.

\paragraph{Bias consideration}
Three kinds of biases may affect the MJDs: option bias, explanation sequence bias and prompt length bias.
\begin{itemize}
    \item Option bias, such as the LLM may prefer the first option A~\citep[e.g.,][]{DBLP:journals/corr/abs-2306-07951,DBLP:conf/iclr/Zheng0M0H24,DBLP:journals/tacl/TjuatjaCWTN24}, could be addressed by shuffling the mapping relationship between \texttt{[A,B,C]} and \texttt{[Entailment, Neutral, Contradiction]}, resulting in $A\binom {3}{3}=6$ permutations. 
    \item Explanation sequence bias, representing the LLM may be affected by the sequence of $m$ input explanations, could be addressed by using the average output of full permutations $A\binom {m}{m}$ as the model’s final answer.
    \item Prompt length bias arises when the LLM may perform differently facing input prompts with different lengths. For in total $m$ explanations, we consider gradually increasing the number of explanations $n$ simultaneously put in one promo (denote as ``n in one''), that contains $C\binom {n}{m}$ combinations. Then we average all the scores from $\sum_{n=1}^{m}C\binom {n}{m}$ combinations to get a length-independent result.
\end{itemize}

\begin{table*}[t]
\centering
\resizebox{\textwidth}{!}{
\begin{tabular}{l|l|c|c|c}
\toprule
\textbf{Dataset Name} & 
\textbf{Number of Instances} & \textbf{Annotations per Instance} & \textbf{Explanations} & \textbf{Valid Overlap} \\ \midrule
MNLI \cite{DBLP:conf/naacl/WilliamsNB18}  & 433K total, 40K multi-label & 1 or 5 & No & 341 \\
ChaosNLI \citep{DBLP:conf/acl/NieWDBWK20} & 1.5K from each of $\alpha$NLI, SNLI, MNLI
& 100 & No & 341 \\ 
VariErr NLI \cite{DBLP:conf/acl/Weber-GenzelPMP24} & 500 & 4  & 1 per label & 341 \\ 
ANLI test \cite{DBLP:conf/acl/NieWDBWK20} & 1K (R1), 1K (R2), 1.2K (R3) & 1 &  Yes (Rationale) & 0 \\ 
\bottomrule
\end{tabular}
}
\caption{NLI Datasets with multiple labels and/or explanation annotations in this paper.}
\label{tab:nli_datasets}
\end{table*}

\paragraph{Datasets}

We experiment on four NLI datasets (see Table~\ref{tab:nli_datasets} for details):

\begin{itemize}
    \item {Chaos NLI}~\cite{DBLP:conf/emnlp/NieZB20} is annotated by 100 crowd workers for capturing human judgment distributions. In this paper, we consider label distributions from Chaos NLI as the ``gold label'' that the LLM approximates.
    \item {VariErr NLI}~\cite{DBLP:conf/acl/Weber-GenzelPMP24} is annotated by 4 experts, who also wrote down their explanations for why they chose. We use those explanations as ``human explanations'' in our paper.
    \item {MNLI}~\cite{DBLP:conf/naacl/WilliamsNB18} is annotated by 5 annotators. Its dev set contains all NLI instances of Chaos NLI and VariErr NLI. We use the valid overlap of these three datasets (341 NLI instanced with human judgment distribution and 4 human explanations for each) as the target datasets in this paper. For the other part of the MNLI subset of Chaos NLI, we divide them into dev and test sets for evaluation.
    \item {ANLI}~\cite{DBLP:conf/acl/NieWDBWK20} is annotated by adversarial human-and-model-in-the-loop procedure, which is not overlap with above datasets. We utilize the test set of ANLI for the out-of-domain evaluation.
\end{itemize}

\begin{table}[t]
    \centering
    \begin{adjustbox}{width={0.35\textwidth},keepaspectratio}%

    \begin{tabular}{l r}
        \toprule 
        \textbf{Hyperparameter} & \textbf{Our Model} \\
        \midrule
        Learning Rate Decay & Linear \\
        Weight Decay & 0.0 \\
        Optimizer & AdamW \\
        Adam $\epsilon$ & 1e-8 \\
        Adam $\beta_{1}$ & 0.9 \\
        Adam $\beta_{2}$ & 0.999 \\
        Warmup Ratio & 0\% \\
        Learning Rate & 2e-5 \\
        Batch size & 4 \\
        Num Epoch & 5\\
        \bottomrule
    \end{tabular}
    \end{adjustbox}

    \caption{Hyperparameter used for fine-tuning BERT and RoBERTa models with soft labels. }
    \label{tab:hyperpara}
\end{table}

\paragraph{Evaluation Protocals}
Following \citet{DBLP:conf/emnlp/Chen0PLKP24}, we evaluated the obtained MJDs on Distribution Comparison, Fine-tuning comparison and Global-Metric Comparison.

For Distribution Comparison, we investigate these distribution differences between
humans and LLMs at the instance level following prior work \citep{DBLP:conf/emnlp/NieZB20,chiang-lee-2023-large, DBLP:conf/emnlp/LeeAT23,baan-etal-2022-stop,DBLP:conf/emnlp/Chen0PLKP24}:  Kullback-Leibler Divergence (KL, \citealt{kullback1951information}), Jensen-Shannon Distance (JSD, \citealt{DBLP:journals/tit/EndresS03}) and Total Variation Distance (TVD, \citealt{DBLP:books/daglib/0018090}). The calculations for all metrics are listed below:

For discrete probability distributions \(P\) and \(Q\):

\begin{equation}
D_{\text{KL}}(P | Q) = \sum_{x \in \mathcal{X}} P(x) \log \frac{P(x)}{Q(x)} ,
\end{equation}

\begin{equation}
D_{\text{JSD}}(P | Q) = \sqrt{\frac{(D_{\text{KL}}(P | M) +  D_{\text{KL}}(Q | M))}{2} } ,
\end{equation}

\begin{equation}
D_{\text{TVD}}(P, Q) = \frac{1}{2} \sum_{x \in \mathcal{X}} |P(x) - Q(x)| ,
\end{equation}

For Fine-tuning Comparison, we investigate how well the resulting MJDs approximate human labels for model training.  
To do so, we leverage the HJD and human-multi-labels from existing datasets and generated MJDs as labels of the overlapped instances in MNLI, VariErr and ChaosNLI, for fine-tuning smaller language models, namely,  BERT \cite{DBLP:conf/naacl/DevlinCLT19} and RoBERTa \cite{DBLP:journals/corr/abs-1907-11692} base.
These models were first fine-tuned on the large single-labelled MNLI dataset to learn the generic NLI task. We then few-shot-tune them on the HJD, label distributions or MJDs above.
To evaluate the resulting classifiers, we split the remaining 1,258 MNLI instances from ChaosNLI that do not overlap with VariErr NLI into the development and test sets. 
We use KL, Cross-Entropy Loss and Weighted F1 scores as evaluation metrics between the outputs of the fine-tuned models and the models trained by ChaosNLI training set HJD. Detailed hyperparameter choices are listed in Table~\ref{tab:hyperpara}. The formula for the weighted F1 score is:

\begin{equation}
\text{Weighted F1} = \frac{1}{N} \sum_{i=1}^{k} w_i \times F1_i .
\end{equation}

\noindent where 

\begin{equation}
\text{F1 Score} = 2 \times \frac{\text{Precision} \times \text{Recall}}{\text{Precision} + \text{Recall}} ,
\end{equation}

\begin{equation}
\text{Precision} = \frac{TP}{TP + FP} ,
\end{equation}
    
\begin{equation}
\text{Recall} = \frac{TP}{TP + FN} ,
\end{equation}

Moreover, for Global-Metric Comparison we further evaluate MJDs against HJD using a global-level measure, distance correlation (D.Corr, \citealt{szekely2007measuring}), to capture the differences between general distributions. The D.Corr between the source dataset $X$ and the target dataset $Y$ is calculated as: 
\begin{equation}
    \operatorname{dCor}^2(X, Y)=\frac{\mathrm{dCov}^2(X, Y)}{\sqrt{\operatorname{dVar}^2(X) \operatorname{dVar}^2(Y)}} .
\end{equation}

\begin{table}[t]
    \centering
    \resizebox{\linewidth}{!}{
    \begin{tabular}{lccc}
    \toprule 
\textbf{LLM} & \textbf{LF} & \textbf{VariErr LG} & \textbf{MNLI LG} \\
\midrule
Llama3   & 11.05\%    & 29.84\%              & 19.77\%           \\
GPT-4o     & 9.78\%     & 32.77\%              & 24.93\%           \\
Mixtral & 6.26\%     & 19.57\%              & 16.07\%            \\      
\bottomrule
    \end{tabular}}
    \caption{Overlap rate of model explanations between \textit{first} and \textit{longest} under different selection settings.}
    \label{tab:stats_firstlong}
\end{table}

\begin{figure*}[t]

        \centering
        \includegraphics[width=\linewidth]{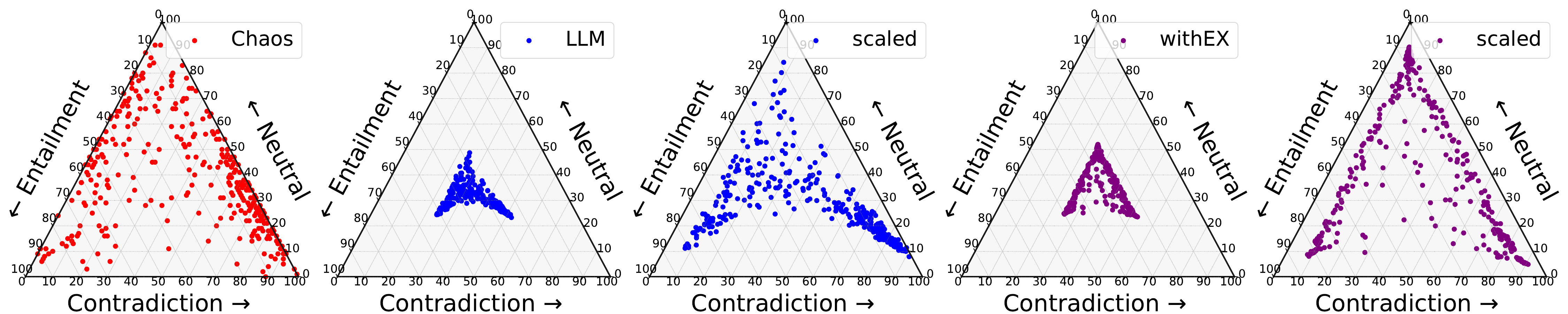}
        \caption{Llama3 with human explanations.}
        \label{fig:visual-main-Llama-process}
\end{figure*}

\begin{figure*}[t]

        \centering
        \includegraphics[width=\linewidth]{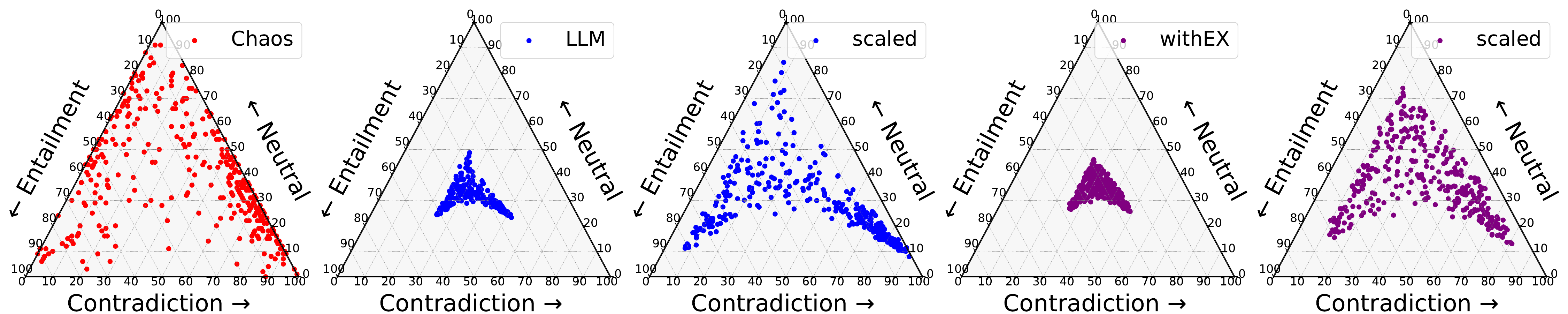}
        \caption{Llama3 with Lable-Free first model explanations.}
        \label{fig:visual-main-diversity-first}
\end{figure*}

\begin{figure*}[t]

        \centering
        \includegraphics[width=\linewidth]{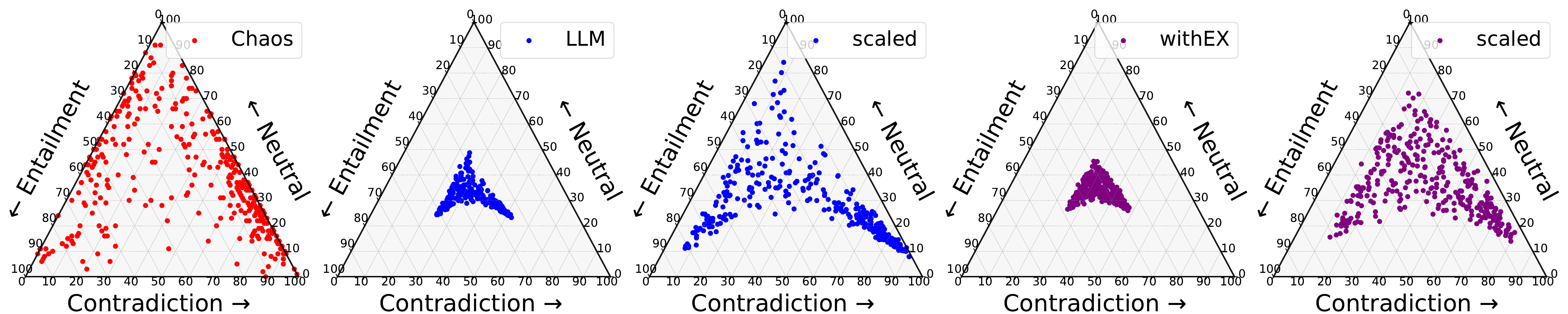}
        \caption{Llama3 with Lable-Free longest model explanations.}
        \label{fig:visual-main-diversity-long}
\end{figure*}

\begin{figure*}[t]

        \centering
        \includegraphics[width=\linewidth]{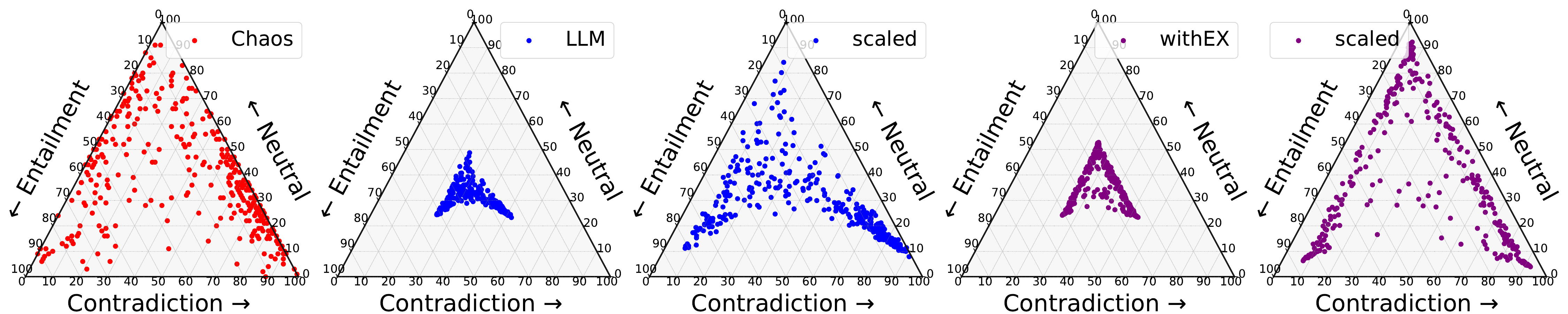}
        \caption{Llama3 with VariErr Label-Guided first model explanations.}
        \label{fig:visual-main-replace-first}
\end{figure*}

\begin{figure*}[t]

        \centering
        \includegraphics[width=\linewidth]{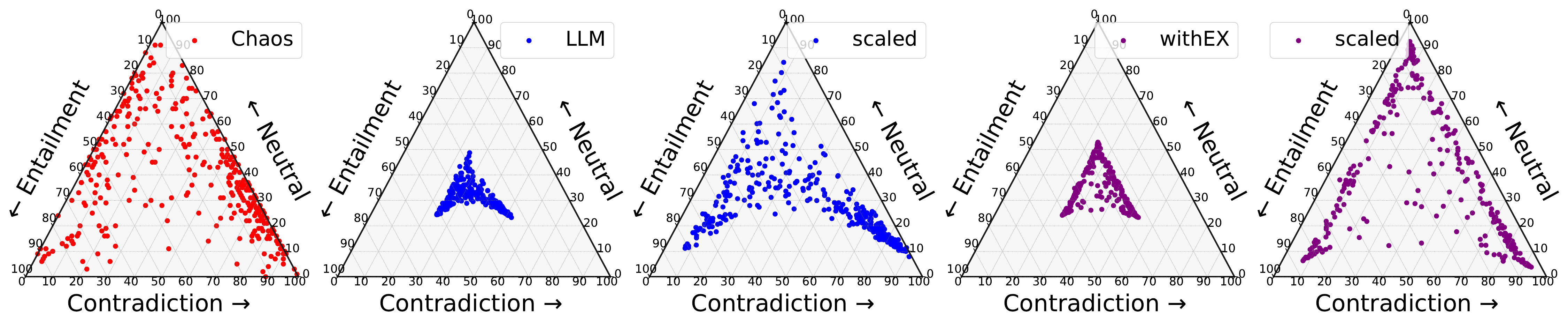}
        \caption{Llama3 with VariErr Label-Guided longest model explanations.}
        \label{fig:visual-main-replace-long}
\end{figure*}

\begin{figure*}[t]

        \centering
        \includegraphics[width=\linewidth]{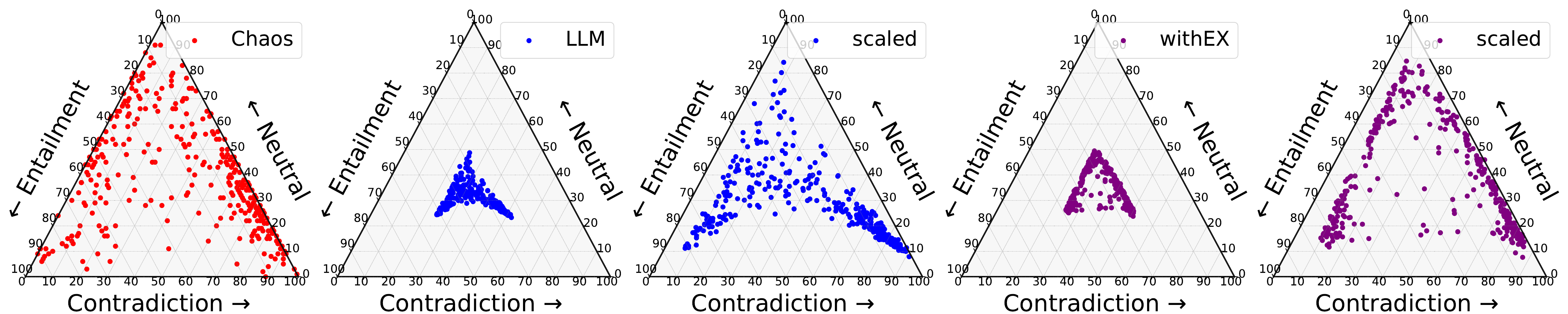}
        \caption{Llama3 with MNLI Label-Guided first model explanations.}
        \label{fig:visual-main-MNLI-first}
\end{figure*}

\begin{figure*}[t]

        \centering
        \includegraphics[width=\linewidth]{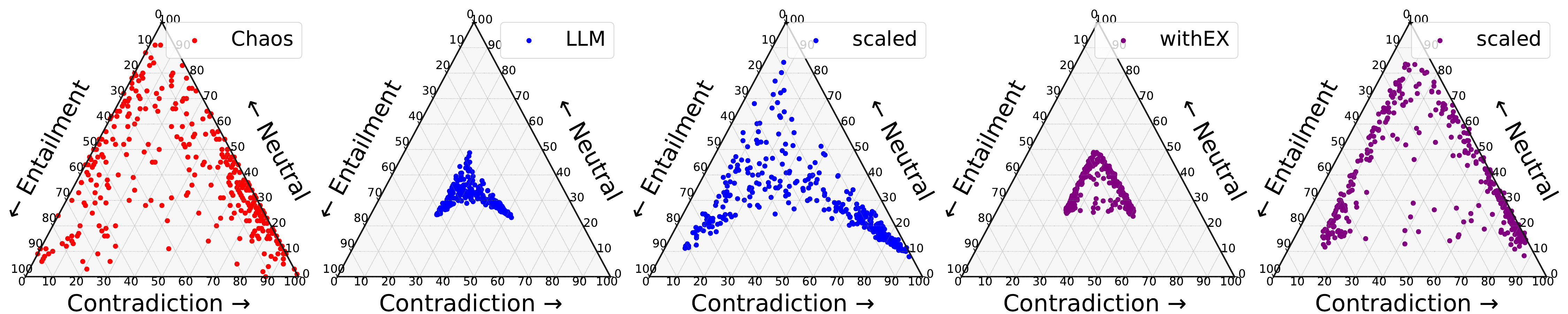}
        \caption{Llama3 with MNLI Label-Guided longest model explanations.}
        \label{fig:visual-main-MNLI-long}
\end{figure*}

\begin{table*}[t]
\centering
\resizebox{\textwidth}{!}{
\begin{tabular}{lccc|ccc|ccc|c}
\toprule
\multicolumn{1}{c}{\multirow{2}{*}{\textbf{Distributions}}} & \multicolumn{3}{c|}{\textbf{Dist. Comparison}} & \multicolumn{3}{c|}{\textbf{BERT Fine-Tuning Comparison(dev/test)}} & \multicolumn{3}{c|}{\textbf{RoBERTa Fine-Tuning Comparison(dev/test)}} &  \textbf{Global Metric}\\ \cmidrule(lr){2-11} 
\multicolumn{1}{c}{}          & {KL $\downarrow$} & {JSD $\downarrow$} & {TVD $\downarrow$}                                       & {KL $\downarrow$} &  {CE Loss $\downarrow$}  & {Weighted F1 $\uparrow$}    & {KL $\downarrow$}  &  {CE Loss $\downarrow$} & {Weighted F1 $\uparrow$} & {D.Corr $\uparrow$}    \\
\midrule
\multicolumn{11}{l}{\textit{Baseline from Human Annotations}}   \\
\midrule
 ChaosNLI HJD     & 0.000 & 0.000 & 0.000 & 0.073 / 0.077 & 0.967 / 0.974 & 0.645 / 0.609 & 0.062 / 0.060 & 0.933 / 0.922 & 0.696 / 0.653 & 1.000 \\
VariErr dist.  & 3.604 & 0.282 & 0.296 & 0.177 / 0.179 & 1.279 / 1.279 & 0.552 / 0.522 & 0.166 / 0.173 & 1.246 / 1.261 & 0.616 / 0.594 & 0.688 \\
MNLI dist.    & 1.242 & 0.281 & 0.295 & 0.104 / 0.100 & 1.062 / 1.042 & 0.569 / 0.555 & 0.101 / 0.093 & 1.052 / 1.020 & 0.625 / 0.607 & 0.795 \\
\midrule
\multicolumn{11}{l}{\textit{Model Judgment Distributions}}   \\
\midrule
Llama3 & 0.259 & 0.262 & 0.284 & 0.099 / 0.101 & 1.045 / 1.044 & 0.516 / 0.487 & 0.094 / 0.096 & 1.030 / 1.031 & 0.545 / 0.522 & 0.689 \\
\multicolumn{11}{l}{{+ human explanations}}   \\
\, 4 in one     & 0.235 & 0.247 & 0.266 & 0.097 / 0.098 & 1.040 / 1.036 & 0.571 / 0.553 & 0.089 / 0.090 & 1.016 / 1.013 & 0.631 / 0.610 & 0.733 \\
\, 3 in one     & 0.235 & 0.248 & 0.266 & 0.098 / 0.099 & 1.041 / 1.038 & 0.580 / 0.560 & 0.090 / 0.091 & 1.018 / 1.016 & 0.640 / 0.623 & 0.757 \\
\, 2 in one     & 0.238 & 0.250 & 0.269 & 0.098 / 0.099 & 1.043 / 1.040 & 0.578 / 0.561 & 0.091 / 0.093 & 1.023 / 1.021 & 0.640 / 0.611 & 0.784 \\
\, 1 in one     & 0.243 & 0.253 & 0.273 & 0.099 / 0.100 & 1.046 / 1.044 & 0.572 / 0.549 & 0.093 / 0.094 & 1.027 / 1.025 & 0.651 / 0.619 & 0.809 \\
\, avg          & 0.238 & 0.250 & 0.269 & 0.098 / 0.099 & 1.043 / 1.039 & 0.575 / 0.556 & 0.091 / 0.092 & 1.021 / 1.019 & 0.641 / 0.616 & 0.771 \\
\midrule
\multicolumn{11}{l}{{+ first model explanations}}   \\
\multicolumn{11}{l}{{\, \, Label-Free}}   \\
\, \, 3 in one     & 0.281 & 0.271 & 0.300 & 0.102 / 0.103 & 1.054 / 1.051 & 0.581 / 0.570 & 0.098 / 0.100 & 1.043 / 1.042 & 0.662 / 0.613 & 0.713 \\
\, \, 2 in one     & 0.292 & 0.276 & 0.308 & 0.105 / 0.106 & 1.063 / 1.060 & 0.544 / 0.538 & 0.102 / 0.104 & 1.056 / 1.055 & 0.599 / 0.593 & 0.748 \\
\, \, 1 in one     & 0.305 & 0.282 & 0.316 & 0.108 / 0.109 & 1.073 / 1.069 & 0.519 / 0.520 & 0.107 / 0.108 & 1.068 / 1.067 & 0.578 / 0.543 & 0.762 \\
\, \, avg          & 0.293 & 0.276 & 0.308 & 0.105 / 0.106 & 1.063 / 1.060 & 0.548 / 0.543 & 0.102 / 0.104 & 1.056 / 1.054 & 0.613 / 0.583 & 0.741 \\
\multicolumn{11}{l}{{\, \, VariErr Label-Guided}}   \\
\, \, 4 in one     & 0.234 & 0.246 & 0.264 & 0.097 / 0.098 & 1.038 / 1.035 & 0.538 / 0.541 & 0.088 / 0.089 & 1.012 / 1.010 & 0.619 / 0.622 & 0.722 \\
\, \, 3 in one     & 0.233 & 0.246 & 0.264 & 0.097 / 0.098 & 1.040 / 1.036 & 0.550 / 0.544 & 0.089 / 0.090 & 1.015 / 1.012 & 0.621 / 0.635 & 0.747 \\
\, \, 2 in one     & 0.235 & 0.248 & 0.267 & 0.098 / 0.099 & 1.042 / 1.038 & 0.564 / 0.546 & 0.089 / 0.091 & 1.017 / 1.015 & 0.636 / 0.632 & 0.768 \\
\, \, 1 in one     & 0.241 & 0.251 & 0.272 & 0.099 / 0.099 & 1.045 / 1.040 & 0.554 / 0.541 & 0.091 / 0.093 & 1.023 / 1.020 & 0.631 / 0.623 & 0.784 \\
\, \, avg          & 0.236 & 0.248 & 0.267 & 0.098 / 0.098 & 1.041 / 1.037 & 0.551 / 0.543 & 0.089 / 0.091 & 1.017 / 1.014 & 0.627 / 0.628 & 0.755 \\
\multicolumn{11}{l}{{\, \, MNLI Label-Guided}}   \\
\, \, 5 in one     & 0.237 & 0.248 & 0.270 & 0.096 / 0.096 & 1.035 / 1.030 & 0.594 / 0.576 & 0.088 / 0.089 & 1.012 / 1.010 & 0.656 / 0.657 & 0.811 \\
\, \, 4 in one     & 0.239 & 0.250 & 0.272 & 0.096 / 0.097 & 1.036 / 1.032 & 0.586 / 0.579 & 0.089 / 0.090 & 1.015 / 1.013 & 0.659 / 0.655 & 0.827 \\
\, \, 3 in one     & 0.242 & 0.251 & 0.275 & 0.096 / 0.097 & 1.037 / 1.033 & 0.593 / 0.583 & 0.090 / 0.091 & 1.018 / 1.016 & 0.663 / 0.654 & 0.842 \\
\, \, 2 in one     & 0.247 & 0.254 & 0.279 & 0.097 / 0.098 & 1.039 / 1.036 & 0.598 / 0.585 & 0.091 / 0.093 & 1.022 / 1.021 & 0.672 / 0.650 & 0.856 \\
\, \, 1 in one     & 0.255 & 0.257 & 0.285 & 0.098 / 0.099 & 1.043 / 1.038 & 0.586 / 0.565 & 0.093 / 0.095 & 1.028 / 1.027 & 0.667 / 0.636 & 0.863 \\
\, \, avg          & 0.244 & 0.252 & 0.276 & 0.097 / 0.097 & 1.038 / 1.034 & 0.591 / 0.577 & 0.090 / 0.092 & 1.019 / 1.017 & 0.663 / 0.650 & 0.840 \\
\midrule
\multicolumn{11}{l}{{+ longest model explanations}}   \\
\multicolumn{11}{l}{{\, \, Label-Free}}   \\
\, \, 3 in one     & 0.285 & 0.274 & 0.303 & 0.103 / 0.105 & 1.058 / 1.056 & 0.550 / 0.558 & 0.100 / 0.102 & 1.049 / 1.048 & 0.615 / 0.595 & 0.714 \\
\, \, 2 in one     & 0.296 & 0.278 & 0.311 & 0.106 / 0.107 & 1.066 / 1.064 & 0.533 / 0.525 & 0.104 / 0.106 & 1.060 / 1.059 & 0.551 / 0.561 & 0.750 \\
\, \, 1 in one     & 0.305 & 0.282 & 0.317 & 0.108 / 0.109 & 1.073 / 1.070 & 0.535 / 0.516 & 0.107 / 0.108 & 1.068 / 1.067 & 0.578 / 0.556 & 0.769 \\
\, \, avg          & 0.295 & 0.278 & 0.310 & 0.106 / 0.107 & 1.066 / 1.063 & 0.539 / 0.533 & 0.103 / 0.105 & 1.059 / 1.058 & 0.581 / 0.571 & 0.744 \\
\multicolumn{11}{l}{{\, \, VariErr Label-Guided}}   \\
\, \, 4 in one     & 0.231 & 0.245 & 0.263 & 0.096 / 0.098 & 1.038 / 1.035 & 0.551 / 0.541 & 0.087 / 0.089 & 1.011 / 1.009 & 0.630 / 0.623 & 0.736 \\
\, \, 3 in one     & 0.231 & 0.245 & 0.263 & 0.097 / 0.098 & 1.039 / 1.036 & 0.562 / 0.542 & 0.088 / 0.090 & 1.013 / 1.012 & 0.632 / 0.622 & 0.754 \\
\, \, 2 in one     & 0.234 & 0.247 & 0.266 & 0.097 / 0.099 & 1.041 / 1.038 & 0.558 / 0.544 & 0.089 / 0.091 & 1.017 / 1.014 & 0.633 / 0.631 & 0.771 \\
\, \, 1 in one     & 0.240 & 0.250 & 0.271 & 0.099 / 0.099 & 1.045 / 1.040 & 0.562 / 0.546 & 0.091 / 0.092 & 1.022 / 1.019 & 0.635 / 0.627 & 0.781 \\
\, \, avg          & 0.234 & 0.247 & 0.266 & 0.097 / 0.098 & 1.041 / 1.037 & 0.558 / 0.544 & 0.089 / 0.091 & 1.016 / 1.014 & 0.633 / 0.626 & 0.760 \\
\multicolumn{11}{l}{{\, \, MNLI Label-Guided}}   \\
\, \, 5 in one     & 0.234 & 0.247 & 0.268 & 0.095 / 0.096 & 1.034 / 1.031 & 0.582 / 0.579 & 0.088 / 0.090 & 1.012 / 1.011 & 0.654 / 0.644 & 0.833 \\
\, \, 4 in one     & 0.237 & 0.249 & 0.271 & 0.096 / 0.097 & 1.035 / 1.032 & 0.591 / 0.580 & 0.089 / 0.091 & 1.015 / 1.014 & 0.651 / 0.646 & 0.843 \\
\, \, 3 in one     & 0.240 & 0.250 & 0.273 & 0.096 / 0.097 & 1.037 / 1.034 & 0.588 / 0.590 & 0.090 / 0.092 & 1.017 / 1.017 & 0.652 / 0.646 & 0.852 \\
\, \, 2 in one     & 0.245 & 0.253 & 0.278 & 0.097 / 0.098 & 1.039 / 1.036 & 0.593 / 0.582 & 0.091 / 0.093 & 1.022 / 1.021 & 0.660 / 0.648 & 0.859 \\
\, \, 1 in one     & 0.255 & 0.257 & 0.285 & 0.098 / 0.099 & 1.043 / 1.039 & 0.592 / 0.568 & 0.093 / 0.095 & 1.028 / 1.027 & 0.668 / 0.639 & 0.859 \\
\, \, avg          & 0.242 & 0.251 & 0.275 & 0.096 / 0.097 & 1.037 / 1.034 & 0.589 / 0.580 & 0.090 / 0.092 & 1.019 / 1.018 & 0.657 / 0.645 & 0.849
\\
\bottomrule
\end{tabular}}
\caption{Main evaluation results for individual runs corresponding to Table~\ref{tab:results-main}. }\label{tab:results-main-raw}
\end{table*}

\subsection{Detailed Main Results}
\label{app:exp1-detail-results}

In this section, we report the detailed results together with \textit{first} and \textit{longest} modes in Table~\ref{tab:results-main-raw} for the main experiment in \S\ref{sec:llm-explanation-approximate-HJD}. Also, we report the statistic information regarding the overlapped model explanations between \textit{first} and \textit{longest} mode in Table~\ref{tab:stats_firstlong}.

In order to mitigate the prompt bias, as illustrated in Appendix~\ref{app:exp1-expset}, we first average the output MJDs in $A\binom {3}{3}=6$ permutations of label-option mappings to avoid ``option bias'', and then divided the MJDs and average them in $A\binom {n}{n}$ permutations (to avoid ``sequence bias'') into ``n in one'' settings where n explanations are fed into LLMs together. We report the scores for ``n in one'' in Table~\ref{tab:results-main-raw} and average them to get the final evaluation results in Table~\ref{tab:results-main} to avoid ``length bias''.

For the ternary visualization, following~\citet{DBLP:conf/emnlp/Chen0PLKP24} we zoom in on the original ternary plot of Llama3 for a clear observation (\textit{scale=3}), since ``shape'' of the ternary distribution is more important as demonstrated by previous work. For Mixtral and GPT-4o we use the original shape of MJDs as they are clear enough. The zooming-in process for Llama3 with human explanations is shown in Figure~\ref{fig:visual-main-Llama-process}. All other zooming-in processes for MJDs (including \textit{first} and \textit{longest} modes) in Table~\ref{tab:results-main} are listed in Figure~\ref{fig:visual-main-diversity-first},~\ref{fig:visual-main-diversity-long},~\ref{fig:visual-main-replace-first},~\ref{fig:visual-main-replace-long},~\ref{fig:visual-main-MNLI-first}, and~\ref{fig:visual-main-MNLI-long}.

\subsection{Detailed Main Results on Mixtral}
\label{app:exp1-detail-results-mixtral}

\begin{figure*}[t]

        \centering
        \includegraphics[width=\linewidth]{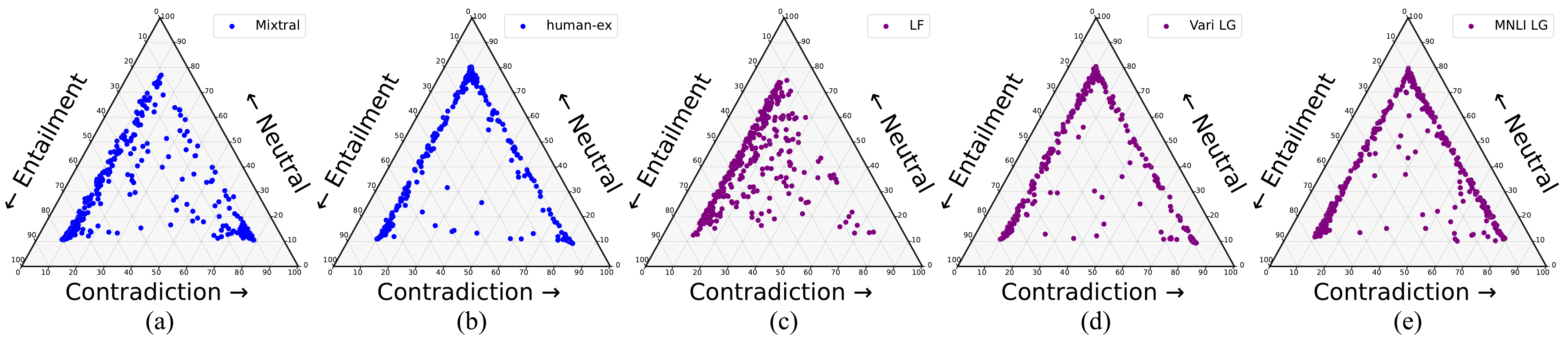}
        \caption{Mixtral Visualization.}
        \label{fig:visual-main-mixtral}
\end{figure*}

Here we report all the results of Mixtral in Table~\ref{tab:results-main-mixtral} under the main experiment settings corresponding to \S\ref{sec:llm-explanation-approximate-HJD}. Also we visualize the MJDs from Mixtral in Figure~\ref{fig:visual-main-mixtral}.

We observe the same findings with previous work~\cite{DBLP:conf/emnlp/Chen0PLKP24} that Mixtral still fails to capture HLV. From Table~\ref{tab:results-main-mixtral}, overall performance is poor, though explanations still provide some benefit. However, unlike other LLMs, Mixtral performs exceptionally poorly in the Label-Free setting. This highlights that Mixtral struggles to effectively capture HLV (human label variation) information from explanations. We can empirically hypothesize that if the Label-Free setting improves performance relative to the original setting, it indicates that explanation information is being effectively utilized. If not, the LLM fails to leverage HLV information from explanations.
The overall conclusion is the same that \textbf{LLM-generated explanations continue to perform similarly to human explanations across all metrics}. 

For the Mixtral visualization in Figure~\ref{fig:visual-main-mixtral},
incorrect predictions are concentrated on the left side, whereas HJD (human judgment distribution) is primarily on the right side.
The second figure (human explanations) and the fourth figure (VariErr Label-Guided explanations) perform similarly.
The Labal-Free setting further worsens performance, with more predictions concentrated on the left side, resulting in poor evaluation scores that align with Table~\ref{tab:results-main-mixtral}. This demonstrates that Mixtral struggles to effectively utilize explanation information.
On MNLI-Label-Guided, explanations slightly help shift some points toward the correct right-side direction, showing marginal improvement.
Overall, Mixtral's performance remains weak, consistent with findings from the \citet{DBLP:conf/emnlp/Chen0PLKP24}.

\begin{table*}[t]
\centering
\resizebox{\textwidth}{!}{
\begin{tabular}{lccc|ccc|ccc|c}
\toprule
\multicolumn{1}{c}{\multirow{2}{*}{\textbf{Distributions}}} & \multicolumn{3}{c|}{\textbf{Dist. Comparison}} & \multicolumn{3}{c|}{\textbf{BERT Fine-Tuning Comparison (dev/test)}} & \multicolumn{3}{c|}{\textbf{RoBERTa Fine-Tuning Comparison (dev/test)}} &  \textbf{Global Metric}\\ \cmidrule(lr){2-11} 
\multicolumn{1}{c}{}          & {KL $\downarrow$} & {JSD $\downarrow$} & {TVD $\downarrow$}                                       & {KL $\downarrow$} &  {CE Loss $\downarrow$}  & {Weighted F1 $\uparrow$}    & {KL $\downarrow$}  &  {CE Loss $\downarrow$} & {Weighted F1 $\uparrow$} & {D.Corr $\uparrow$}    \\
\midrule
\multicolumn{11}{l}{\textit{Baseline from Human Annotations}}   \\
\midrule
 ChaosNLI HJD          & 0.000          & 0.000          & 0.000          & 0.073          / 0.077          & 0.967          / 0.974          & 0.645          / 0.609          & 0.062          / 0.060          & 0.933          / 0.922          & 0.696          / 0.653          & 1.000          \\
VariErr distribution       & 3.604          & 0.282          & 0.296          & 0.177          / 0.179          & 1.279          / 1.279          & 0.552          / 0.522          & 0.166          / 0.173          & 1.246          / 1.261          & 0.616          / 0.594          & 0.688          \\
MNLI distribution         & 1.242          & 0.281          & 0.295          & 0.104          / 0.100          & 1.062          / 1.042          & 0.569          / 0.555          & 0.101          / 0.093          & 1.052          / 1.020          & 0.625          / 0.607          & 0.795          \\
\midrule
\multicolumn{11}{l}{\textit{Model Judgment Distributions}}   \\
\midrule
Mixtral       & 0.437 & 0.293 & 0.324 & 0.131 / 0.129 & 1.140 / 1.130 & 0.427 / 0.432 & 0.121 / 0.125 & 1.112 / 1.118 & 0.497 / 0.472 & 0.593 \\
\rowcolor{green!20}
+ human explanations & 0.239 & 0.225 & 0.257 & 0.121 / 0.109 & 1.112 / 1.075 & 0.525 / 0.519 & \textbf{0.086} / \textbf{0.085} & 1.007 / 0.998 & 0.575 / 0.557 & 0.656 \\
\multicolumn{11}{l}{{+ model explanations}} \\
\, \, Label-Free            & 0.361 & 0.299 & 0.343 & 0.233 / 0.222 & 1.447 / 1.407 & 0.298 / 0.296 & 0.241 / 0.237 & 1.472 / 1.452 & 0.274 / 0.302 & 0.483 \\
\rowcolor{green!20}
\, \, VariErr Label-Guided              & 0.238 & 0.224 & 0.255 & 0.108 / 0.097 & 1.073 / 1.032 & 0.519 / 0.513 & 0.091 / 0.089 & 1.021 / 1.010 & 0.569 / 0.557 & 0.642 \\
\, \, MNLI Label-Guided                 & \textbf{0.237} & \textbf{0.223} & \textbf{0.253} & \textbf{0.097} / \textbf{0.095} & \textbf{1.041} / \textbf{1.028} & \textbf{0.530} / \textbf{0.533} & 0.086 / 0.085 & \textbf{1.006} / \textbf{0.997} & \textbf{0.575} / \textbf{0.559} & \textbf{0.726}
\\
\bottomrule
\end{tabular}}
\caption{Evaluation results for Mixtral. }\label{tab:results-main-mixtral}
\end{table*}

\section{Detailed ANLI Test Results}
\label{app:anli}

In this section we report the detailed results in Table~\ref{tab:results-main-ANLI-raw} for evaluation of the ANLI test set. Also we listed the scores of Mixtral on the ANLI test in Table~\ref{tab:results-main-ANLI-Mixtral}. All the performances of MJDs in the ANLI test set are aligned with those in the main experiments in Table~\ref{tab:results-main}, which demonstrate the generalization capability of our method on OOD tasks. 

\begin{table}[t]
\centering
\resizebox{\linewidth}{!}{
\begin{tabular}{lccc|ccc}
\toprule
\multicolumn{1}{c}{\multirow{2}{*}{\textbf{Classifiers}}} & \multicolumn{3}{c|}{\textbf{BERT FT Test}} & \multicolumn{3}{c}{\textbf{RoBERTa FT Test}} \\ \cmidrule(lr){2-7} 
\multicolumn{1}{c}{}              & {R1 $\uparrow$} &  {R2 $\uparrow$}  & {R3 $\uparrow$}   & {R1 $\uparrow$} &  {R2 $\uparrow$}  & {R3 $\uparrow$}   \\
\midrule
\multicolumn{7}{l}{\textit{Classifiers without distribution training}}   \\
\midrule
Out-of-the-box LM  & 0.170 & 0.176 & 0.197 & 0.167 & 0.167 & 0.168 \\
MNLI-FT-LM   & 0.220 & 0.269 & 0.293 & 0.292 & 0.262 & 0.257 \\
\midrule
\multicolumn{7}{l}{\textit{Classifier trained on label distributions}}   \\
\midrule
ChaosNLI HJD     & 0.268 & 0.289 & 0.332 & 0.357 & 0.331 & 0.338 \\
VariErr dist.  & 0.302 & 0.259 & 0.319 & 0.402 & 0.311 & 0.321 \\
MNLI dist.    & 0.229 & 0.260 & 0.279 & 0.317 & 0.275 & 0.281 \\
\midrule
\multicolumn{7}{l}{\textit{Classifier trained on MJDs}}   \\
\midrule
Llama origin & 0.246 & 0.276 & 0.306 & 0.304 & 0.297 & 0.304 \\
\multicolumn{7}{l}{{+ human explanations}}   \\
\, \, 4 in one     & 0.294 & 0.287 & 0.349 & 0.403 & 0.335 & 0.349 \\
\, \, 3 in one     & 0.304 & 0.288 & 0.349 & 0.406 & 0.325 & 0.344 \\
\, \, 2 in one     & 0.301 & 0.291 & 0.351 & 0.407 & 0.335 & 0.344 \\
\, \, 1 in one     & 0.298 & 0.291 & 0.348 & 0.397 & 0.325 & 0.338 \\
\, \, avg          & 0.296 & 0.289 & 0.349 & 0.400 & 0.330 & 0.344 \\
\midrule
\multicolumn{7}{l}{{+ first model explanations}}   \\
\multicolumn{7}{l}{{\, \, Label-Free}}   \\
\, \, 3 in one     & 0.300 & 0.299 & 0.353 & 0.356 & 0.283 & 0.340 \\
\, \, 2 in one     & 0.293 & 0.293 & 0.327 & 0.296 & 0.244 & 0.321 \\
\, \, 1 in one     & 0.276 & 0.276 & 0.297 & 0.257 & 0.224 & 0.284 \\
\, \, avg          & 0.288 & 0.288 & 0.325 & 0.307 & 0.254 & 0.312 \\
\multicolumn{7}{l}{{\, \, VariErr Label-Guided}}   \\
\, \, 4 in one     & 0.294 & 0.269 & 0.345 & 0.412 & 0.335 & 0.312 \\
\, \, 3 in one     & 0.296 & 0.271 & 0.353 & 0.407 & 0.341 & 0.321 \\
\, \, 2 in one     & 0.303 & 0.280 & 0.358 & 0.403 & 0.336 & 0.312 \\
\, \, 1 in one     & 0.318 & 0.293 & 0.346 & 0.391 & 0.310 & 0.313 \\
\, \, avg          & 0.306 & 0.281 & 0.345 & 0.402 & 0.322 & 0.312 \\
\multicolumn{7}{l}{{\, \, MNLI Label-Guided}}   \\
\, \, 5 in one     & 0.294 & 0.281 & 0.323 & 0.354 & 0.300 & 0.311 \\
\, \, 4 in one     & 0.286 & 0.290 & 0.324 & 0.351 & 0.290 & 0.314 \\
\, \, 3 in one     & 0.280 & 0.284 & 0.323 & 0.346 & 0.286 & 0.321 \\
\, \, 2 in one     & 0.272 & 0.285 & 0.316 & 0.342 & 0.280 & 0.314 \\
\, \, 1 in one     & 0.271 & 0.285 & 0.304 & 0.318 & 0.269 & 0.286 \\
\, \, avg          & 0.282 & 0.283 & 0.314 & 0.336 & 0.284 & 0.298 \\
\midrule
\multicolumn{7}{l}{{+ longest model explanations}}   \\
\multicolumn{7}{l}{{\, \, Label-Free}}   \\
\, \, 3 in one     & 0.308 & 0.302 & 0.352 & 0.340 & 0.288 & 0.352 \\
\, \, 2 in one     & 0.281 & 0.279 & 0.312 & 0.286 & 0.250 & 0.321 \\
\, \, 1 in one     & 0.277 & 0.288 & 0.304 & 0.288 & 0.235 & 0.295 \\
\, \, avg          & 0.292 & 0.295 & 0.328 & 0.314 & 0.262 & 0.323 \\
\multicolumn{7}{l}{{\, \, VariErr Label-Guided}}   \\
\, \, 4 in one     & 0.298 & 0.284 & 0.351 & 0.417 & 0.335 & 0.325 \\
\, \, 3 in one     & 0.293 & 0.283 & 0.350 & 0.416 & 0.337 & 0.314 \\
\, \, 2 in one     & 0.295 & 0.281 & 0.350 & 0.419 & 0.338 & 0.316 \\
\, \, 1 in one     & 0.312 & 0.287 & 0.348 & 0.405 & 0.313 & 0.313 \\
\, \, avg          & 0.305 & 0.285 & 0.349 & 0.411 & 0.324 & 0.319 \\
\multicolumn{7}{l}{{\, \, MNLI Label-Guided}}   \\
\, \, 5 in one     & 0.288 & 0.281 & 0.330 & 0.354 & 0.301 & 0.327 \\
\, \, 4 in one     & 0.283 & 0.277 & 0.332 & 0.351 & 0.297 & 0.336 \\
\, \, 3 in one     & 0.282 & 0.283 & 0.328 & 0.349 & 0.289 & 0.334 \\
\, \, 2 in one     & 0.278 & 0.285 & 0.323 & 0.343 & 0.282 & 0.319 \\
\, \, 1 in one     & 0.279 & 0.286 & 0.312 & 0.325 & 0.272 & 0.287 \\
\, \, avg          & 0.284 & 0.283 & 0.321 & 0.339 & 0.287 & 0.307
\\
\bottomrule
\end{tabular}}
\caption{ANLI test results for individual runs. }\label{tab:results-main-ANLI-raw}
\end{table}

\begin{table}[t]
\centering
\resizebox{\linewidth}{!}{
\begin{tabular}{lccc|ccc}
\toprule
\multicolumn{1}{c}{\multirow{2}{*}{\textbf{Classifiers}}} & \multicolumn{3}{c|}{\textbf{BERT FT Test}} & \multicolumn{3}{c}{\textbf{RoBERTa FT Test}} \\ \cmidrule(lr){2-7} 
\multicolumn{1}{c}{}              & {R1 $\uparrow$} &  {R2 $\uparrow$}  & {R3 $\uparrow$}   & {R1 $\uparrow$} &  {R2 $\uparrow$}  & {R3 $\uparrow$}   \\
\midrule
\multicolumn{7}{l}{\textit{Classifiers without distribution training}}   \\
\midrule
Out-of-the-box LM        & 0.170          & 0.176          & 0.197          & 0.167          & 0.167          & 0.168          \\
MNLI-FT-LM         & 0.220          & 0.269          & 0.293          & 0.292          & 0.262          & 0.257          \\
\midrule
\multicolumn{7}{l}{\textit{Classifiers trained on label distributions}}   \\
\midrule
ChaosNLI HJD           & 0.268          & 0.289          & 0.332          & 0.357          & \textbf{0.331} & 0.338          \\
\rowcolor{green!20}
VariErr distribution      & 0.302          & 0.259          & 0.319          & 0.402          & 0.311          & 0.321          \\
\rowcolor{yellow!20}
MNLI distribution          & 0.229          & 0.260          & 0.279          & 0.317          & 0.275          & 0.281          \\
\midrule
\multicolumn{7}{l}{\textit{Classifiers trained on MJDs}}   \\
\midrule
Mixtral   & 0.242 & 0.252 & 0.246 & 0.230 & 0.240 & 0.243 \\
\rowcolor{green!20}
+ human explanations      & \textbf{0.344} & 0.280 & \textbf{0.320} & 0.361 & \textbf{0.292} & \textbf{0.300} \\
\multicolumn{7}{l}{{+ model explanations}}   \\
\, \, Label-Free   & 0.252 & 0.255 & 0.255 & 0.242 & 0.248 & 0.243 \\
\rowcolor{green!20}
\, \, VariErr Label-Guided   & 0.340 & \textbf{0.287} & 0.317 & \textbf{0.362} & 0.289 & 0.296 \\
\rowcolor{yellow!20}
\, \, MNLI Label-Guided    & 0.275 & 0.273 & 0.303 & 0.329 & 0.280 & 0.292
\\
\bottomrule
\end{tabular}}
\caption{ANLI test results for Mixtral. }\label{tab:results-main-ANLI-Mixtral}
\end{table}
\section{Detailed Ablation Results}
\label{app:ablation}

\begin{figure*}[t]

        \centering
        \includegraphics[width=\linewidth]{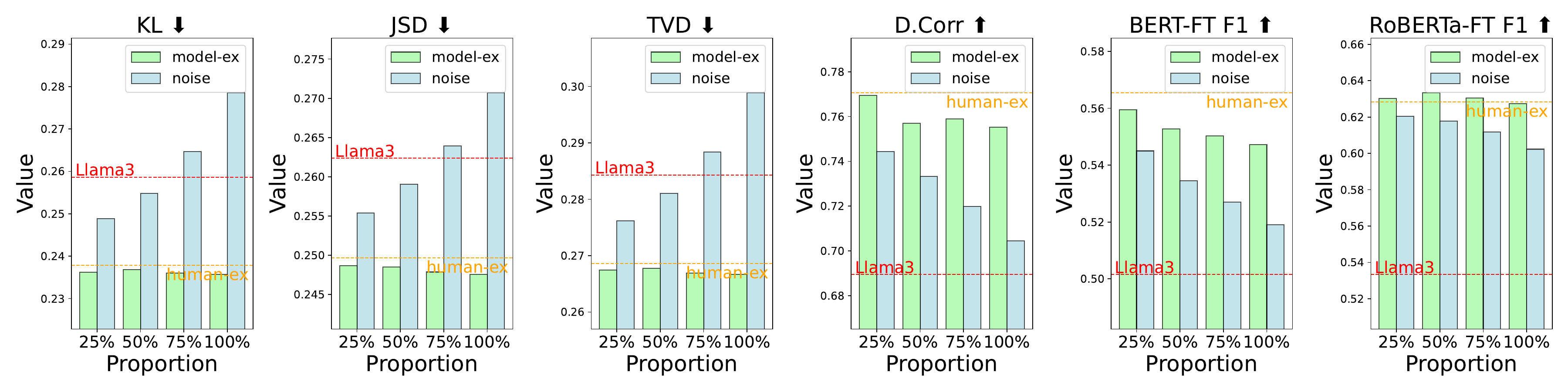}
        \caption{Results for ablation study (Llama3) on gradually replacing first model/noise explanations.}
        \label{fig:ablation-bar-replace-first}
\end{figure*}

\begin{figure}[t]

        \centering
        \includegraphics[width=\linewidth]{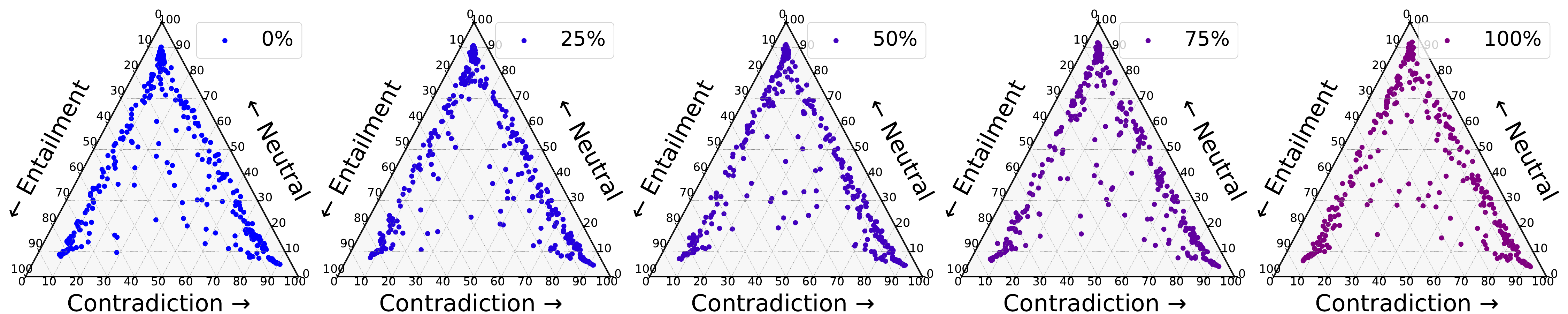}
        \caption{Visualization for gradually replacing first model explanations (Llama3 Scaled by 3).}
        \label{fig:ablation-visual-first-auto}
\end{figure}

\begin{figure}[t]

        \centering
        \includegraphics[width=\linewidth]{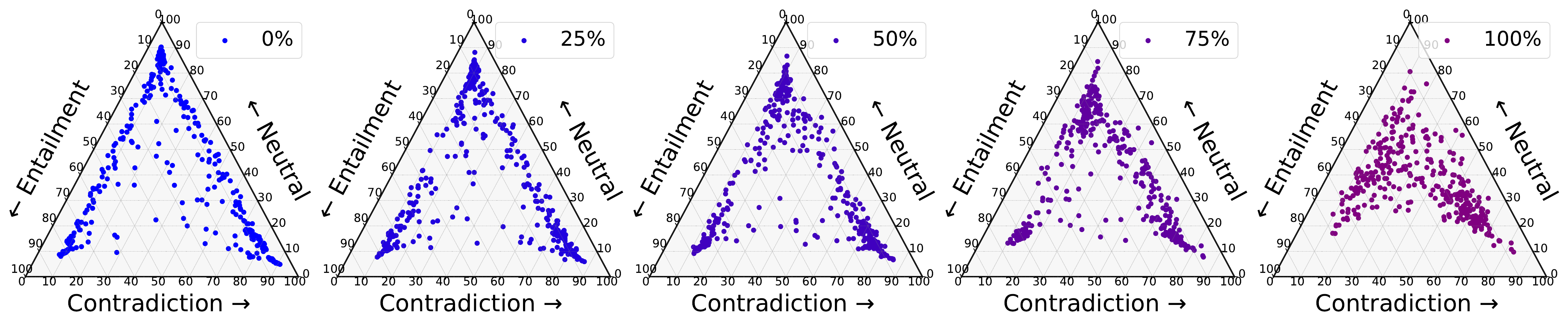}
        \caption{Visualization gradually replacing first noise explanations (Llama3 Scaled by 3).}
        \label{fig:ablation-visual-first-noise}
\end{figure}

\begin{figure}[t]

        \centering
        \includegraphics[width=\linewidth]{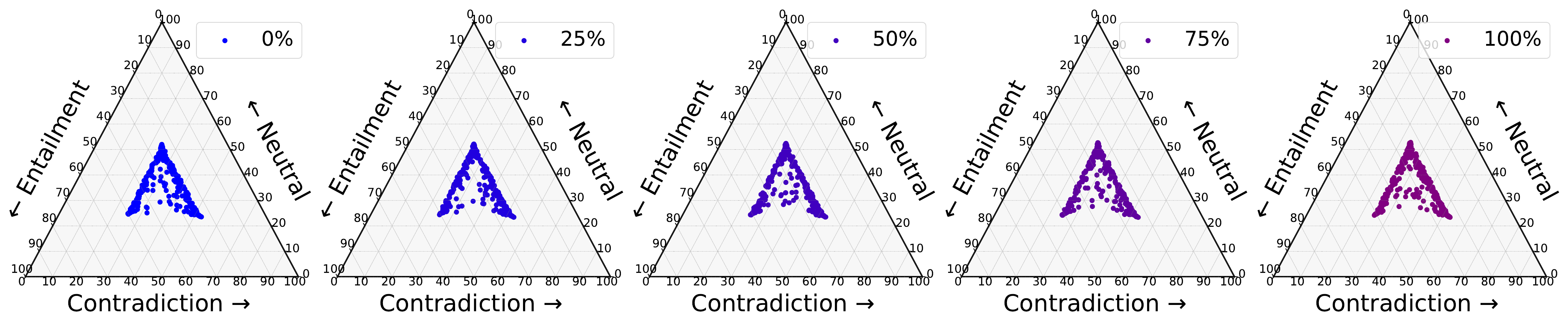}
        \caption{Visualization for gradually replacing first model explanations (Llama3).}
        \label{fig:ablation-origin-visual-first-auto}
\end{figure}

\begin{figure}[t]

        \centering
        \includegraphics[width=\linewidth]{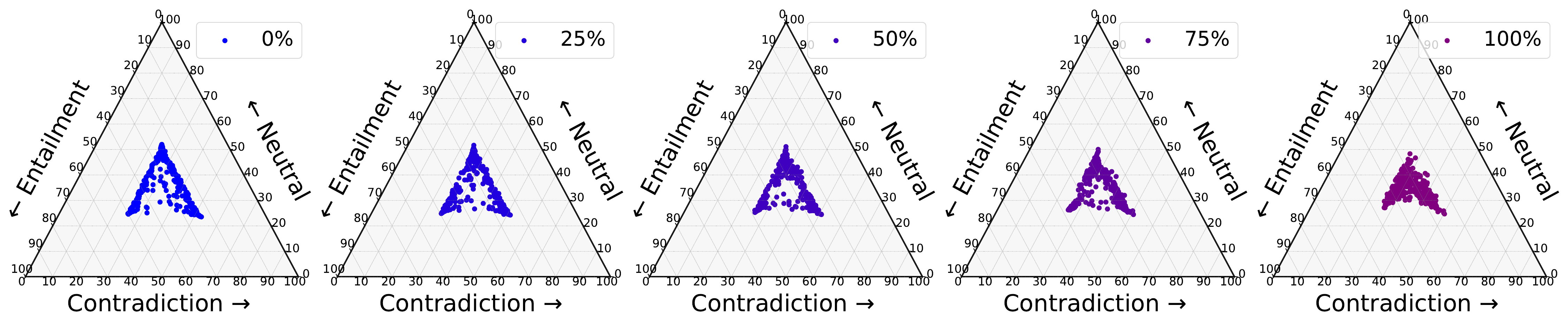}
        \caption{Visualization gradually replacing first noise explanations (Llama3).}
        \label{fig:ablation-origin-visual-first-noise}
\end{figure}

\begin{figure}[t]

        \centering
        \includegraphics[width=\linewidth]{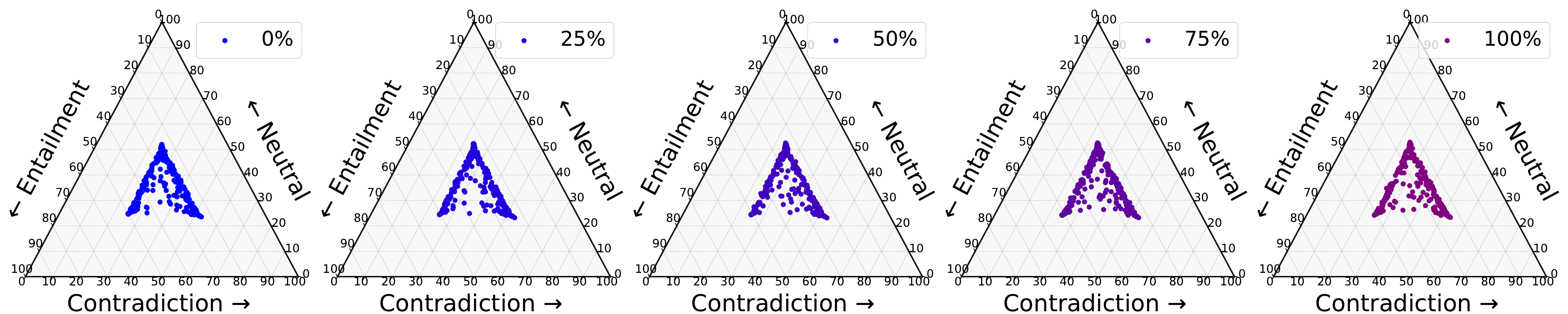}
        \caption{Visualization for gradually replacing longest model explanations (Llama3).}
        \label{fig:ablation-origin-visual-long-auto}
\end{figure}

\begin{figure}[t]

        \centering
        \includegraphics[width=\linewidth]{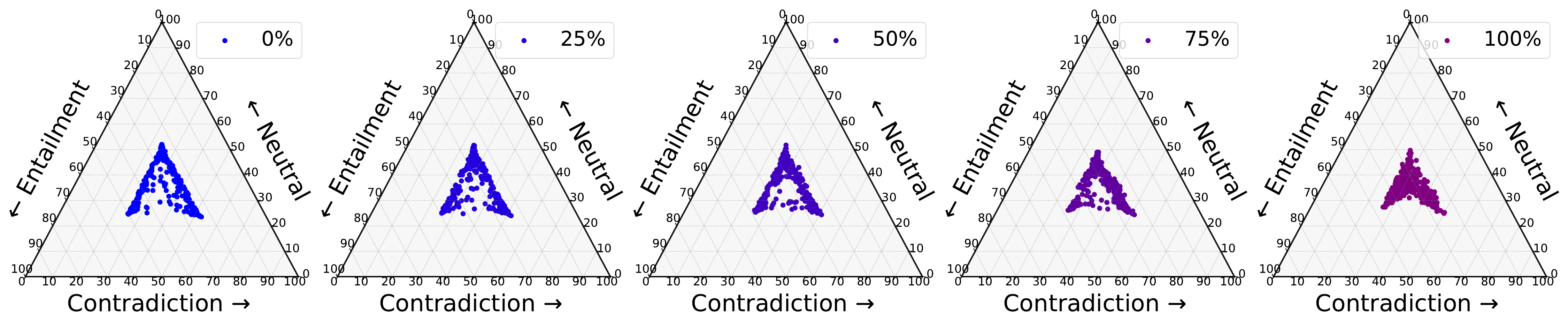}
        \caption{Visualization for gradually replacing longest noise explanations (Llama3).}
        \label{fig:ablation-origin-visual-long-noise}
\end{figure}

In this section, we report the detailed scores (for both \textit{first} and \textit{longest} modes) of the Figure~\ref{fig:ablation-bar-replace-long} in Table~\ref{tab:results-main-ANLI-raw}. We also plot the bar figure for the ablation study on the \textit{first} mode, as shown in Figure~\ref{fig:ablation-bar-replace-first}.
For the visualizations, the original MJDs and zooming-in MJDs of replaced model/noise explanations under both \textit{first} and \textit{longest} modes are plotted in Figure~\ref{fig:ablation-origin-visual-first-auto},~\ref{fig:ablation-origin-visual-first-noise},~\ref{fig:ablation-origin-visual-long-auto},~\ref{fig:ablation-origin-visual-long-noise},~\ref{fig:ablation-visual-first-auto}, and~\ref{fig:ablation-visual-first-noise}.
Results from a human-explanation-centric view in \textit{first} mode are also listed in Table~\ref{tab:results-3-ablation-centric}.
All the findings remain basically the same for \textit{first} and \textit{longest} modes.

\paragraph{Linguistic Similarities}
Even though our experiments so far show that model explanations are comparable to human explanations in helping LLMs approximate HJD, we next wonder to what degree they are similar linguistically.
Following \citet{giulianelli-etal-2023-comes}, we adopt their method to measure similarity across multiple references (in our case, explanations) along three axes (lexical, syntactic, semantic).
The similarities between model explanations and corresponding human explanations are listed in Table~\ref{tab:results-ablation-similarity-parallel}. 
Our observation mirrors theirs in that model explanations generated by LLMs are moderately different from human explanations regarding lexicon, syntax, and semantics.
Nevertheless, despite these linguistic differences, LLMs still found a way to obtain comparable information for modeling HJD.
label distribution. 
We leave this matter for future investigation. 

\begin{table*}[t]
\centering
\resizebox{\textwidth}{!}{
\begin{tabular}{lccc|ccc|ccc|c}
\toprule
\multicolumn{1}{c}{\multirow{2}{*}{\textbf{Distributions}}} & \multicolumn{3}{c|}{\textbf{Dist. Comparison}} & \multicolumn{3}{c|}{\textbf{BERT Fine-Tuning Comparison(dev/test)}} & \multicolumn{3}{c|}{\textbf{RoBERTa Fine-Tuning Comparison(dev/test)}} &  \textbf{Global Metric}\\ \cmidrule(lr){2-11} 
\multicolumn{1}{c}{}          & {KL $\downarrow$} & {JSD $\downarrow$} & {TVD $\downarrow$}                                       & {KL $\downarrow$} &  {CE Loss $\downarrow$}  & {Weighted F1 $\uparrow$}    & {KL $\downarrow$}  &  {CE Loss $\downarrow$} & {Weighted F1 $\uparrow$} & {D.Corr $\uparrow$}    \\
\midrule
\multicolumn{11}{l}{\textit{Baseline from Human Annotations}}   \\
\midrule
 ChaosNLI HJD          & 0.000 & 0.000 & 0.000 & 0.073 / 0.077 & 0.967   / 0.974   & 0.645       / 0.609       & 0.062 / 0.060 & 0.933   / 0.922   & 0.696       / 0.653       & 1.000  \\
VariErr distribution      & 3.604 & 0.282 & 0.296 & 0.177 / 0.179 & 1.279   / 1.279   & 0.552       / 0.522       & 0.166 / 0.173 & 1.246   / 1.261   & 0.616       / 0.594       & 0.688  \\
MNLI distribution         & 1.242 & 0.281 & 0.295 & 0.104 / 0.100 & 1.062   / 1.042   & 0.569       / 0.555       & 0.101 / 0.093 & 1.052   / 1.020   & 0.625       / 0.607       & 0.795  \\
\midrule
\multicolumn{11}{l}{\textit{Model Judgment Distributions}}   \\
\midrule
Llama3       & 0.259 & 0.262 & 0.284 & 0.099 / 0.101 & 1.045   / 1.044   & 0.516       / 0.487       & 0.094 / 0.096 & 1.030   / 1.031   & 0.545       / 0.522       & 0.689  \\
+ human explanations & 0.238 & 0.250 & 0.269 & 0.098 / 0.099 & 1.043   / 1.039   & 0.575       / 0.556       & 0.091 / 0.092 & 1.021   / 1.019   & 0.641       / 0.616       & 0.771  \\
\multicolumn{11}{l}{{+ replace first model explanations}} \\
\, \, Label-Free 100\%             & 0.293 & 0.276 & 0.308 & 0.105 / 0.106 & 1.063   / 1.060   & 0.548       / 0.543       & 0.102 / 0.104 & 1.056   / 1.054   & 0.613       / 0.583       & 0.741  \\
\, \, noise       & 0.292 & 0.276 & 0.308 & 0.105 / 0.106 & 1.063   / 1.060   & 0.510       / 0.504       & 0.102 / 0.104 & 1.056   / 1.055   & 0.549       / 0.543       & 0.702  \\
\, \, VariErr Label-Guided 25\%   & 0.236 & 0.249 & 0.267 & 0.098 / 0.099 & 1.042   / 1.038   & 0.572       / 0.547       & 0.090 / 0.091 & 1.019   / 1.016   & 0.639       / 0.622       & 0.769  \\
\, \, noise             & 0.249 & 0.255 & 0.276 & 0.100 / 0.101 & 1.048   / 1.045   & 0.551       / 0.539       & 0.094 / 0.095 & 1.030   / 1.029   & 0.628       / 0.613       & 0.744  \\
\, \, VariErr Label-Guided 50\%    & 0.237 & 0.248 & 0.268 & 0.098 / 0.098 & 1.041   / 1.038   & 0.559       / 0.546       & 0.089 / 0.091 & 1.017   / 1.015   & 0.639       / 0.628       & 0.757  \\
\, \, noise              & 0.255 & 0.259 & 0.281 & 0.099 / 0.101 & 1.047   / 1.044   & 0.546       / 0.523       & 0.094 / 0.096 & 1.032   / 1.031   & 0.625       / 0.610       & 0.733  \\
\, \, VariErr Label-Guided 75\%    & 0.236 & 0.248 & 0.267 & 0.098 / 0.098 & 1.041   / 1.037   & 0.557       / 0.544       & 0.090 / 0.091 & 1.018   / 1.015   & 0.634       / 0.628       & 0.759  \\
\, \, noise              & 0.265 & 0.264 & 0.288 & 0.101 / 0.102 & 1.050   / 1.048   & 0.533       / 0.521       & 0.096 / 0.098 & 1.037   / 1.036   & 0.622       / 0.601       & 0.720  \\
\, \, VariErr Label-Guided 100\%    & 0.236 & 0.248 & 0.267 & 0.098 / 0.098 & 1.041   / 1.037   & 0.551       / 0.543       & 0.089 / 0.091 & 1.017   / 1.014   & 0.627       / 0.628       & 0.755  \\
\, \, noise              & 0.279 & 0.271 & 0.299 & 0.102 / 0.103 & 1.055   / 1.052   & 0.525       / 0.513       & 0.099 / 0.101 & 1.046   / 1.045   & 0.612       / 0.592       & 0.705  \\
\midrule
\multicolumn{11}{l}{{+ replace longest model explanations}} \\
\, \, Label-Free 100\%              & 0.295 & 0.278 & 0.310 & 0.106 / 0.107 & 1.066   / 1.063   & 0.539       / 0.533       & 0.103 / 0.105 & 1.059   / 1.058   & 0.581       / 0.571       & 0.744  \\
\, \, noise         & 0.288 & 0.275 & 0.306 & 0.104 / 0.105 & 1.061   / 1.058   & 0.516       / 0.515       & 0.101 / 0.103 & 1.052   / 1.053   & 0.558       / 0.552       & 0.709  \\
\, \, VariErr Label-Guided 25\%     & 0.236 & 0.249 & 0.267 & 0.098 / 0.099 & 1.042   / 1.038   & 0.574       / 0.551       & 0.090 / 0.091 & 1.019   / 1.016   & 0.641       / 0.627       & 0.772  \\
\, \, noise              & 0.248 & 0.255 & 0.275 & 0.100 / 0.101 & 1.048   / 1.044   & 0.561       / 0.540       & 0.094 / 0.095 & 1.029   / 1.028   & 0.631       / 0.618       & 0.745  \\
\, \, VariErr Label-Guided 50\%     & 0.236 & 0.248 & 0.267 & 0.097 / 0.099 & 1.041   / 1.038   & 0.571       / 0.543       & 0.090 / 0.091 & 1.017   / 1.016   & 0.639       / 0.627       & 0.757  \\
\, \, noise              & 0.253 & 0.258 & 0.280 & 0.099 / 0.101 & 1.046   / 1.044   & 0.546       / 0.531       & 0.094 / 0.096 & 1.031   / 1.030   & 0.620       / 0.616       & 0.735  \\
\, \, VariErr Label-Guided 75\%     & 0.235 & 0.248 & 0.267 & 0.097 / 0.099 & 1.041   / 1.038   & 0.564       / 0.545       & 0.090 / 0.091 & 1.018   / 1.016   & 0.643       / 0.622       & 0.760  \\
\, \, noise              & 0.261 & 0.262 & 0.286 & 0.100 / 0.101 & 1.049   / 1.046   & 0.535       / 0.521       & 0.095 / 0.097 & 1.034   / 1.034   & 0.620       / 0.609       & 0.723  \\
\, \, VariErr Label-Guided 100\%     & 0.234 & 0.247 & 0.266 & 0.097 / 0.098 & 1.041   / 1.037   & 0.558       / 0.544       & 0.089 / 0.091 & 1.016   / 1.014   & 0.633       / 0.626       & 0.760  \\
\, \, noise              & 0.274 & 0.269 & 0.296 & 0.101 / 0.103 & 1.053   / 1.050   & 0.526       / 0.511       & 0.098 / 0.100 & 1.042   / 1.042   & 0.608       / 0.599       & 0.709 \\
\bottomrule
\end{tabular}}
\caption{Detailed scores for the ablation study. }\label{tab:results-ablation}
\end{table*}

\begin{table}[t]
\centering
\resizebox{\linewidth}{!}{
\begin{tabular}{lccc|c}
\toprule
\multicolumn{1}{c}{\multirow{2}{*}{\textbf{Distributions}}} & \multicolumn{3}{c|}{\textbf{Dist. Comparison}} &  \textbf{Global Metric}\\ \cmidrule(lr){2-5} 
\multicolumn{1}{c}{}          & {KL $\downarrow$} & {JSD $\downarrow$} & {TVD $\downarrow$}                                      & {D.Corr $\uparrow$}    \\
\midrule
VariErr distributions       & 6.628          & 0.357          & 0.352          & 0.907          \\
Llama3 MJD       & 0.029          & 0.068          & 0.088          & 0.691          \\
\rowcolor{green!20}
+ human explanations & 0.000          & 0.000          & 0.000          & 1.000          \\
\midrule
\multicolumn{5}{l}{{+ replace first model explanations}} \\
\, \, Label-Free 100\%              & 0.023          & 0.066          & 0.087          & 0.645          \\
\, \, VariErr Label-Guided 25\%    & 0.002          & 0.013          & 0.017          & 0.970          \\
\, \, VariErr Label-Guided 50\%    & 0.003          & 0.018          & 0.023          & 0.955          \\
\, \, VariErr Label-Guided 75\%    & 0.003          & 0.020          & 0.026          & 0.945          \\
\, \, VariErr Label-Guided 100\%    & 0.005          & 0.023          & 0.029          & 0.930          \\
\midrule
\multicolumn{5}{l}{{+ replace longest model explanations}} \\
\, \, Label-Free 100\%                & 0.024          & 0.067          & 0.088          & 0.647          \\
\, \, VariErr Label-Guided 25\%     & \textbf{0.001} & \textbf{0.012} & \textbf{0.015} & \textbf{0.977} \\
\, \, VariErr Label-Guided 50\%     & 0.003          & 0.017          & 0.022          & 0.959          \\
\, \, VariErr Label-Guided 75\%     & 0.003          & 0.019          & 0.024          & 0.950          \\
\, \, VariErr Label-Guided 100\%     & 0.004          & 0.021          & 0.027          & 0.939        
\\
\bottomrule
\end{tabular}}
\caption{All results from a human-explanation-centric view.
All MJDs are compared to the MJD in the green row.}\label{tab:results-3-ablation-centric}
\end{table}

\begin{table*}[t]
\centering
\resizebox{\textwidth}{!}{
\begin{tabular}{lccc|ccc|cc|c}
\toprule
\multicolumn{1}{c}{\multirow{2}{*}{\textbf{Datasets}}} & \multicolumn{3}{c|}{\textbf{Lexical}} & \multicolumn{3}{c|}{\textbf{Syntactic}} & \multicolumn{2}{c|}{\textbf{Semantic}} &  \textbf{AVG}\\ \cmidrule(lr){2-10} 
\multicolumn{1}{c}{}          & {n = 1 $\uparrow$} & {n = 2 $\uparrow$} & {n = 3 $\uparrow$}                                     & {n = 1 $\uparrow$} & {n = 2 $\uparrow$} & {n = 3 $\uparrow$}         & {Cos. $\uparrow$}  &  {Euc. $\uparrow$}  & {AVG $\uparrow$}    \\
\midrule
human explanations & 1.000 & 1.000 & 1.000 & 1.000 & 1.000 & 1.000 & 1.000 & 1.000 & 1.000 \\
\midrule
\multicolumn{10}{l}{{replace first model explanations}}   \\
\, 25\%  & 0.773 & 0.673 & 0.647 & 0.929 & 0.769 & 0.686 & 0.841 & 0.828 & 0.801 \\
\, 50\%  & 0.603 & 0.432 & 0.389 & 0.873 & 0.598 & 0.457 & 0.712 & 0.698 & 0.651 \\
\, 75\%  & 0.459 & 0.239 & 0.184 & 0.832 & 0.468 & 0.277 & 0.608 & 0.593 & 0.526 \\
\, 100\% & 0.358 & 0.103 & 0.041 & 0.805 & 0.377 & 0.149 & 0.529 & 0.519 & 0.439 \\
\midrule
\multicolumn{10}{l}{{replace longest model explanations}}   \\
\, 25\%   & 0.758 & 0.658 & 0.635 & 0.926 & 0.761 & 0.674 & 0.824 & 0.819 & 0.789 \\
\, 50\%   & 0.581 & 0.416 & 0.378 & 0.873 & 0.592 & 0.447 & 0.691 & 0.690 & 0.635 \\
\, 75\%   & 0.438 & 0.222 & 0.173 & 0.832 & 0.462 & 0.267 & 0.581 & 0.584 & 0.511 \\
\, 100\%   & 0.335 & 0.087 & 0.033 & 0.807 & 0.369 & 0.141 & 0.501 & 0.510 & 0.422
\\
\bottomrule
\end{tabular}}
\caption{Linguistic similarity results for the ablation study. All sets of explanations are parallelly calculated the similarity with human explanations on Lexical, Syntactic and Semantic levels folloing~\citet{giulianelli-etal-2023-comes}. }\label{tab:results-ablation-similarity-parallel}
\end{table*}
\section{Detailed Results for Explanation Selection}
\label{app:validation}

We report the complete comparison in Table~\ref{tab:results-validation-all} for explanation selection strategy based on LLM/human preference, including \textit{first}, \textit{longest} modes based on LLM preference, as well as \textit{preferred} and \textit{unpreferred} modes based on our annotator's preference. All the detailed scores are in Table~\ref{tab:results-validation-raw}. The \textit{unpreferred} explanation set achieves the best performance.

Table~\ref{tab:results-similarity-centric-main} reports the variability check among explanations used in the main experiments. MNLI-guided explanations from Llama3 have the best variability, which is consistent with the results in Table~\ref{tab:results-main}.
However, variability cannot be directly linked to the main results in all circumstances. 
For example, Label-Free explanations are naturally guided by diverse labels, which leads to better variability.
Under the same human label guidance, \textit{variability} can correctly reflect explanations' HLV evaluation level. Further exploration of variability as a reliable indicator for evaluating model explanation is an interesting possible future direction.

\begin{table*}[t]
\centering
\resizebox{\textwidth}{!}{
\begin{tabular}{lccc|ccc|ccc|c}
\toprule
\multicolumn{1}{c}{\multirow{2}{*}{\textbf{Distributions}}} & \multicolumn{3}{c|}{\textbf{Dist. Comparison}} & \multicolumn{3}{c|}{\textbf{BERT Fine-Tuning Comparison(dev/test)}} & \multicolumn{3}{c|}{\textbf{RoBERTa Fine-Tuning Comparison(dev/test)}} &  \textbf{Global Metric}\\ \cmidrule(lr){2-11} 
\multicolumn{1}{c}{}          & {KL $\downarrow$} & {JSD $\downarrow$} & {TVD $\downarrow$}                                       & {KL $\downarrow$} &  {CE Loss $\downarrow$}  & {Weighted F1 $\uparrow$}    & {KL $\downarrow$}  &  {CE Loss $\downarrow$} & {Weighted F1 $\uparrow$} & {D.Corr $\uparrow$}    \\
\midrule
\multicolumn{11}{l}{\textit{Baseline from Human Annotations}}   \\
\midrule
 ChaosNLI HJD              & 0.000          & 0.000          & 0.000          & 0.081          / 0.083          & 0.993          / 0.992          & 0.643          / 0.597          & 0.065          / 0.065          & 0.944          / 0.936          & 0.691          / 0.652          & 1.000          \\
VariErr distribution         & 4.254          & 0.313          & 0.320          & 0.193          / 0.197          & 1.329          / 1.333          & 0.563          / 0.535          & 0.214          / 0.222          & 1.391          / 1.407          & 0.585          / 0.555          & 0.661          \\
MNLI distribution             & 1.215          & 0.281          & 0.290          & 0.105          / 0.103          & 1.064          / 1.051          & 0.553          / 0.540          & 0.092          / 0.086          & 1.024          / 0.999          & 0.615          / 0.604          & 0.743          \\
\midrule
\multicolumn{11}{l}{\textit{Model Judgment Distributions}}   \\
\midrule
Llama3          & 0.258          & 0.261          & 0.286          & 0.092          / 0.093          & 1.024          / 1.020          & 0.514          / 0.471          & 0.092          / 0.095          & 1.025          / 1.026          & 0.531          / 0.512          & 0.684          \\
+ human explanations    & 0.240          & 0.249          & 0.275          & 0.090          / 0.090          & 1.017          / 1.011          & 0.594          / 0.567          & 0.089          / 0.091          & 1.014          / 1.015          & 0.618          / 0.597          & 0.750          \\
\midrule
\multicolumn{11}{l}{{+ replace \textit{first} model explanations}}   \\
\, 50\%  & 0.238          & 0.247          & 0.273          & 0.089          / 0.089          & 1.017          / 1.010          & 0.585          / 0.568          & 0.089          / 0.091          & 1.014          / 1.015          & 0.620          / 0.597          & 0.758          \\
\, 75\%  & 0.237          & 0.246          & 0.272          & 0.090          / 0.090          & 1.018          / 1.011          & 0.577          / 0.565          & 0.088          / 0.091          & 1.013          / 1.014          & 0.620          / 0.586          & 0.759          \\
\, 100\% & 0.237          & {0.246} & {0.271} & 0.089          / 0.090          & 1.017          / 1.011          & 0.581          / 0.566          & 0.088          / 0.090          & 1.013          / 1.014          & 0.617          / 0.586          & 0.755          \\
\multicolumn{11}{l}{{+ replace \textit{longest} model explanations}}   \\
\, 50\%   & 0.238          & 0.247          & 0.273          & {0.089} / 0.089          & {1.016} / 1.009          & 0.586          / 0.566          & 0.088          / 0.091          & 1.013          / 1.014          & 0.618          / 0.600          & 0.749          \\
\, 75\%   & 0.239          & 0.247          & 0.273          & 0.090          / 0.090          & 1.017          / 1.011          & 0.581          / 0.565          & 0.088          / 0.091          & 1.013          / 1.014          & 0.618          / 0.594          & 0.744          \\
\, 100\%  & 0.238          & 0.246          & 0.272          & 0.089          / 0.089          & 1.017          / 1.011          & 0.581          / 0.565          & 0.088          / 0.091          & 1.013          / 1.014          & 0.616          / 0.587          & 0.745          \\
\midrule
\multicolumn{11}{l}{{+ replace \textit{preferred} model explanations}}   \\
\, greedy 75.75\%        & 0.241          & 0.248          & 0.274          & 0.089          / 0.090          & 1.017          / 1.011          & 0.584          / 0.569          & 0.088          / 0.090          & 1.013          / 1.013          & 0.619          / 0.594          & 0.733          \\
\, representative 55.25\%      & 0.240          & 0.248          & 0.274          & 0.089          / 0.090          & 1.016          / 1.011          & 0.587          / 0.567          & 0.088          / 0.091          & 1.013          / 1.014          & 0.619          / 0.597          & 0.739          \\
\multicolumn{11}{l}{{+ replace \textit{unpreferred} model explanations}}   \\
\, greedy 68.5\%         & 0.239          & 0.247          & 0.273          & \textbf{0.089}          / \textbf{0.089} & \textbf{1.016}          / \textbf{1.009} & \textbf{0.589} / \textbf{0.571} & \textbf{0.087} / 0.090          & \textbf{1.011} / 1.012          & \textbf{0.623} / 0.599          & 0.752          \\
\, representative 63.25\%      & \textbf{0.237} & \textbf{0.246}          & \textbf{0.271}          & 0.089          / 0.089          & 1.016          / 1.010          & 0.584          / 0.566          & 0.088          / \textbf{0.090} & 1.011          / \textbf{1.012} & 0.621          / \textbf{0.607} & \textbf{0.761} 

\\
\bottomrule
\end{tabular}}
\caption{All the results on 100 validated NLI instances for explanation selection strategy including LLM/human preference. }\label{tab:results-validation-all}
\end{table*}

\begin{table*}[t]
\centering
\resizebox{0.9\textwidth}{!}{
\begin{tabular}{lccc|ccc|ccc|c}
\toprule
\multicolumn{1}{c}{\multirow{2}{*}{\textbf{Distributions}}} & \multicolumn{3}{c|}{\textbf{Dist. Comparison}} & \multicolumn{3}{c|}{\textbf{BERT Fine-Tuning Comparison(dev/test)}} & \multicolumn{3}{c|}{\textbf{RoBERTa Fine-Tuning Comparison(dev/test)}} &  \textbf{Global Metric}\\ \cmidrule(lr){2-11} 
\multicolumn{1}{c}{}          & {KL $\downarrow$} & {JSD $\downarrow$} & {TVD $\downarrow$}                                       & {KL $\downarrow$} &  {CE Loss $\downarrow$}  & {Weighted F1 $\uparrow$}    & {KL $\downarrow$}  &  {CE Loss $\downarrow$} & {Weighted F1 $\uparrow$} & {D.Corr $\uparrow$}    \\
\midrule
\multicolumn{11}{l}{\textit{Baseline from Human Annotations}}   \\
\midrule
 ChaosNLI HJD    & 0.000 & 0.000 & 0.000 & 0.081 / 0.083 & 0.993 / 0.992 & 0.643 / 0.597 & 0.065 / 0.065 & 0.944 / 0.936 & 0.691 / 0.652 & 1.000 \\
VariErr dist.  & 4.254 & 0.313 & 0.320 & 0.193 / 0.197 & 1.329 / 1.333 & 0.563 / 0.535 & 0.214 / 0.222 & 1.391 / 1.407 & 0.585 / 0.555 & 0.661 \\
MNLI dist.    & 1.215 & 0.281 & 0.290 & 0.105 / 0.103 & 1.064 / 1.051 & 0.553 / 0.540 & 0.092 / 0.086 & 1.024 / 0.999 & 0.615 / 0.604 & 0.743 \\
\midrule
\multicolumn{11}{l}{\textit{Model Judgment Distributions}}   \\
\midrule
Llama3 & 0.258 & 0.261 & 0.286 & 0.092 / 0.093 & 1.024 / 1.020 & 0.514 / 0.471 & 0.092 / 0.095 & 1.025 / 1.026 & 0.531 / 0.512 & 0.684 \\
\multicolumn{11}{l}{{+ human explanations}}   \\
\, \, 4 in one     & 0.240 & 0.247 & 0.273 & 0.088 / 0.088 & 1.013 / 1.006 & 0.589 / 0.566 & 0.087 / 0.090 & 1.011 / 1.011 & 0.607 / 0.591 & 0.703 \\
\, \, 3 in one     & 0.239 & 0.247 & 0.273 & 0.089 / 0.089 & 1.015 / 1.009 & 0.599 / 0.566 & 0.088 / 0.090 & 1.012 / 1.013 & 0.613 / 0.598 & 0.732 \\
\, \, 2 in one     & 0.239 & 0.248 & 0.274 & 0.090 / 0.090 & 1.018 / 1.012 & 0.596 / 0.569 & 0.089 / 0.091 & 1.015 / 1.016 & 0.629 / 0.604 & 0.769 \\
\, \, 1 in one     & 0.244 & 0.252 & 0.279 & 0.092 / 0.092 & 1.024 / 1.018 & 0.593 / 0.567 & 0.091 / 0.093 & 1.020 / 1.021 & 0.622 / 0.596 & 0.795 \\
\, \, avg          & 0.240 & 0.249 & 0.275 & 0.090 / 0.090 & 1.017 / 1.011 & 0.594 / 0.567 & 0.089 / 0.091 & 1.014 / 1.015 & 0.618 / 0.597 & 0.750 \\
\midrule
\multicolumn{11}{l}{{+ replace \textit{first} model explanations}}   \\
\multicolumn{11}{l}{{\, 50\%}}   \\
\, \, 4 in one     & 0.236 & 0.245 & 0.271 & 0.088 / 0.088 & 1.013 / 1.006 & 0.587 / 0.571 & 0.087 / 0.089 & 1.010 / 1.010 & 0.616 / 0.598 & 0.720 \\
\, \, 3 in one     & 0.234 & 0.245 & 0.270 & 0.088 / 0.088 & 1.013 / 1.007 & 0.590 / 0.567 & 0.087 / 0.090 & 1.011 / 1.012 & 0.619 / 0.595 & 0.751 \\
\, \, 2 in one     & 0.237 & 0.247 & 0.272 & 0.089 / 0.090 & 1.017 / 1.011 & 0.586 / 0.570 & 0.089 / 0.091 & 1.015 / 1.016 & 0.621 / 0.605 & 0.772 \\
\, \, 1 in one     & 0.244 & 0.251 & 0.279 & 0.092 / 0.092 & 1.024 / 1.017 & 0.578 / 0.566 & 0.091 / 0.093 & 1.022 / 1.022 & 0.624 / 0.592 & 0.791 \\
\, \, avg          & 0.238 & 0.247 & 0.273 & 0.089 / 0.089 & 1.017 / 1.010 & 0.585 / 0.568 & 0.089 / 0.091 & 1.014 / 1.015 & 0.620 / 0.597 & 0.758 \\
\multicolumn{11}{l}{{\, 75\%}}   \\
\, \, 4 in one     & 0.236 & 0.244 & 0.269 & 0.088 / 0.089 & 1.014 / 1.008 & 0.574 / 0.567 & 0.086 / 0.089 & 1.007 / 1.008 & 0.615 / 0.580 & 0.721 \\
\, \, 3 in one     & 0.234 & 0.244 & 0.269 & 0.089 / 0.089 & 1.015 / 1.009 & 0.580 / 0.565 & 0.087 / 0.089 & 1.009 / 1.010 & 0.615 / 0.586 & 0.752 \\
\, \, 2 in one     & 0.236 & 0.246 & 0.272 & 0.090 / 0.090 & 1.018 / 1.012 & 0.580 / 0.569 & 0.088 / 0.091 & 1.014 / 1.014 & 0.625 / 0.595 & 0.773 \\
\, \, 1 in one     & 0.243 & 0.250 & 0.278 & 0.092 / 0.092 & 1.024 / 1.017 & 0.573 / 0.561 & 0.091 / 0.093 & 1.022 / 1.022 & 0.627 / 0.581 & 0.788 \\
\, \, avg          & 0.237 & 0.246 & 0.272 & 0.090 / 0.090 & 1.018 / 1.011 & 0.577 / 0.565 & 0.088 / 0.091 & 1.013 / 1.014 & 0.620 / 0.586 & 0.759 \\
\multicolumn{11}{l}{{\, 100\%}}   \\
\, \, 4 in one     & 0.237 & 0.243 & 0.268 & 0.088 / 0.088 & 1.013 / 1.007 & 0.582 / 0.565 & 0.087 / 0.089 & 1.009 / 1.009 & 0.608 / 0.580 & 0.718 \\
\, \, 3 in one     & 0.234 & 0.244 & 0.268 & 0.088 / 0.089 & 1.014 / 1.008 & 0.585 / 0.567 & 0.087 / 0.089 & 1.009 / 1.010 & 0.614 / 0.590 & 0.750 \\
\, \, 2 in one     & 0.235 & 0.246 & 0.271 & 0.089 / 0.090 & 1.017 / 1.011 & 0.586 / 0.569 & 0.088 / 0.091 & 1.013 / 1.014 & 0.623 / 0.589 & 0.772 \\
\, \, 1 in one     & 0.243 & 0.250 & 0.278 & 0.092 / 0.092 & 1.023 / 1.017 & 0.571 / 0.563 & 0.091 / 0.093 & 1.021 / 1.021 & 0.623 / 0.583 & 0.781 \\
\, \, avg          & 0.237 & 0.246 & 0.271 & 0.089 / 0.090 & 1.017 / 1.011 & 0.581 / 0.566 & 0.088 / 0.090 & 1.013 / 1.014 & 0.617 / 0.586 & 0.755 \\
\midrule
\multicolumn{11}{l}{{+ replace \textit{longest} model explanations}}   \\
\multicolumn{11}{l}{{\, 50\%}}   \\
\, \, 4 in one     & 0.237 & 0.245 & 0.270 & 0.087 / 0.087 & 1.011 / 1.004 & 0.590 / 0.570 & 0.086 / 0.089 & 1.007 / 1.008 & 0.620 / 0.597 & 0.707 \\
\, \, 3 in one     & 0.236 & 0.245 & 0.270 & 0.088 / 0.088 & 1.013 / 1.007 & 0.590 / 0.564 & 0.087 / 0.089 & 1.009 / 1.010 & 0.610 / 0.605 & 0.737 \\
\, \, 2 in one     & 0.237 & 0.247 & 0.272 & 0.089 / 0.089 & 1.016 / 1.010 & 0.589 / 0.569 & 0.088 / 0.091 & 1.013 / 1.014 & 0.621 / 0.605 & 0.765 \\
\, \, 1 in one     & 0.244 & 0.251 & 0.278 & 0.092 / 0.092 & 1.023 / 1.017 & 0.573 / 0.563 & 0.091 / 0.093 & 1.021 / 1.022 & 0.622 / 0.592 & 0.786 \\
\, \, avg          & 0.238 & 0.247 & 0.273 & 0.089 / 0.089 & 1.016 / 1.009 & 0.586 / 0.566 & 0.088 / 0.091 & 1.013 / 1.014 & 0.618 / 0.600 & 0.749 \\
\multicolumn{11}{l}{{\, 75\%}}   \\
\, \, 4 in one     & 0.238 & 0.245 & 0.270 & 0.088 / 0.088 & 1.013 / 1.007 & 0.586 / 0.563 & 0.086 / 0.089 & 1.007 / 1.008 & 0.620 / 0.595 & 0.703 \\
\, \, 3 in one     & 0.236 & 0.245 & 0.270 & 0.089 / 0.089 & 1.015 / 1.008 & 0.587 / 0.569 & 0.087 / 0.089 & 1.009 / 1.010 & 0.617 / 0.599 & 0.732 \\
\, \, 2 in one     & 0.238 & 0.247 & 0.273 & 0.090 / 0.090 & 1.018 / 1.012 & 0.582 / 0.569 & 0.088 / 0.091 & 1.014 / 1.014 & 0.614 / 0.597 & 0.761 \\
\, \, 1 in one     & 0.244 & 0.251 & 0.279 & 0.092 / 0.092 & 1.024 / 1.017 & 0.568 / 0.558 & 0.091 / 0.093 & 1.022 / 1.022 & 0.622 / 0.586 & 0.781 \\
\, \, avg          & 0.239 & 0.247 & 0.273 & 0.090 / 0.090 & 1.017 / 1.011 & 0.581 / 0.565 & 0.088 / 0.091 & 1.013 / 1.014 & 0.618 / 0.594 & 0.744 \\
\multicolumn{11}{l}{{\, 100\%}}   \\
\, \, 4 in one     & 0.237 & 0.244 & 0.269 & 0.088 / 0.088 & 1.013 / 1.007 & 0.586 / 0.568 & 0.086 / 0.089 & 1.008 / 1.009 & 0.613 / 0.589 & 0.709 \\
\, \, 3 in one     & 0.235 & 0.244 & 0.269 & 0.088 / 0.089 & 1.014 / 1.008 & 0.587 / 0.566 & 0.087 / 0.089 & 1.009 / 1.010 & 0.615 / 0.589 & 0.737 \\
\, \, 2 in one     & 0.237 & 0.246 & 0.272 & 0.089 / 0.090 & 1.017 / 1.011 & 0.587 / 0.566 & 0.088 / 0.091 & 1.013 / 1.014 & 0.614 / 0.590 & 0.762 \\
\, \, 1 in one     & 0.244 & 0.250 & 0.278 & 0.092 / 0.091 & 1.023 / 1.017 & 0.566 / 0.559 & 0.091 / 0.093 & 1.021 / 1.021 & 0.622 / 0.579 & 0.774 \\
\, \, avg          & 0.238 & 0.246 & 0.272 & 0.089 / 0.089 & 1.017 / 1.011 & 0.581 / 0.565 & 0.088 / 0.091 & 1.013 / 1.014 & 0.616 / 0.587 & 0.745 \\
\midrule
\multicolumn{11}{l}{{+ replace \textit{aligned} model explanations}}   \\
\multicolumn{11}{l}{{\, greedy 75.75\%}}   \\
\, \, 4 in one     & 0.240 & 0.246 & 0.272 & 0.088 / 0.088 & 1.012 / 1.006 & 0.590 / 0.566 & 0.087 / 0.089 & 1.009 / 1.009 & 0.615 / 0.593 & 0.692 \\
\, \, 3 in one     & 0.239 & 0.246 & 0.272 & 0.088 / 0.089 & 1.013 / 1.008 & 0.591 / 0.575 & 0.087 / 0.090 & 1.011 / 1.011 & 0.611 / 0.590 & 0.719 \\
\, \, 2 in one     & 0.239 & 0.247 & 0.274 & 0.089 / 0.090 & 1.017 / 1.012 & 0.586 / 0.573 & 0.088 / 0.091 & 1.014 / 1.014 & 0.620 / 0.598 & 0.747 \\
\, \, 1 in one     & 0.244 & 0.250 & 0.278 & 0.092 / 0.092 & 1.024 / 1.018 & 0.568 / 0.564 & 0.090 / 0.092 & 1.020 / 1.020 & 0.633 / 0.595 & 0.774 \\
\, \, avg          & 0.241 & 0.248 & 0.274 & 0.089 / 0.090 & 1.017 / 1.011 & 0.584 / 0.569 & 0.088 / 0.090 & 1.013 / 1.013 & 0.619 / 0.594 & 0.733 \\
\multicolumn{11}{l}{{\, representative 55.25\%}}   \\
\, \, 4 in one     & 0.239 & 0.246 & 0.272 & 0.088 / 0.088 & 1.012 / 1.006 & 0.599 / 0.570 & 0.087 / 0.089 & 1.009 / 1.010 & 0.616 / 0.591 & 0.698 \\
\, \, 3 in one     & 0.237 & 0.246 & 0.272 & 0.088 / 0.089 & 1.013 / 1.008 & 0.595 / 0.568 & 0.087 / 0.090 & 1.010 / 1.011 & 0.609 / 0.603 & 0.730 \\
\, \, 2 in one     & 0.239 & 0.248 & 0.274 & 0.089 / 0.090 & 1.017 / 1.011 & 0.587 / 0.567 & 0.088 / 0.091 & 1.013 / 1.014 & 0.617 / 0.605 & 0.752 \\
\, \, 1 in one     & 0.244 & 0.251 & 0.278 & 0.091 / 0.092 & 1.023 / 1.017 & 0.567 / 0.561 & 0.090 / 0.093 & 1.020 / 1.020 & 0.635 / 0.589 & 0.778 \\
\, \, avg          & 0.240 & 0.248 & 0.274 & 0.089 / 0.090 & 1.016 / 1.011 & 0.587 / 0.567 & 0.088 / 0.091 & 1.013 / 1.014 & 0.619 / 0.597 & 0.739 \\
\midrule
\multicolumn{11}{l}{{+ replace \textit{aligned} model explanations}}   \\
\multicolumn{11}{l}{{\, greedy 68.5\% }}   \\
\, \, 4 in one     & 0.237 & 0.244 & 0.270 & 0.088 / 0.088 & 1.012 / 1.005 & 0.595 / 0.565 & 0.086 / 0.088 & 1.006 / 1.007 & 0.622 / 0.600 & 0.712 \\
\, \, 3 in one     & 0.235 & 0.245 & 0.270 & 0.088 / 0.088 & 1.013 / 1.007 & 0.588 / 0.571 & 0.086 / 0.089 & 1.007 / 1.009 & 0.624 / 0.609 & 0.742 \\
\, \, 2 in one     & 0.238 & 0.247 & 0.273 & 0.089 / 0.089 & 1.016 / 1.010 & 0.591 / 0.574 & 0.088 / 0.090 & 1.011 / 1.013 & 0.624 / 0.606 & 0.768 \\
\, \, 1 in one     & 0.246 & 0.251 & 0.280 & 0.091 / 0.091 & 1.022 / 1.015 & 0.583 / 0.573 & 0.091 / 0.093 & 1.021 / 1.021 & 0.622 / 0.580 & 0.787 \\
\, \, avg          & 0.239 & 0.247 & 0.273 & 0.089 / 0.089 & 1.016 / 1.009 & 0.589 / 0.571 & 0.087 / 0.090 & 1.011 / 1.012 & 0.623 / 0.599 & 0.752 \\
\multicolumn{11}{l}{{\, representative 63.25\%}}   \\
\, \, 4 in one     & 0.235 & 0.244 & 0.268 & 0.088 / 0.088 & 1.012 / 1.006 & 0.587 / 0.560 & 0.086 / 0.088 & 1.006 / 1.006 & 0.622 / 0.605 & 0.721 \\
\, \, 3 in one     & 0.233 & 0.244 & 0.268 & 0.088 / 0.088 & 1.013 / 1.007 & 0.586 / 0.567 & 0.086 / 0.089 & 1.008 / 1.009 & 0.625 / 0.613 & 0.753 \\
\, \, 2 in one     & 0.236 & 0.247 & 0.272 & 0.089 / 0.089 & 1.016 / 1.010 & 0.588 / 0.573 & 0.088 / 0.090 & 1.012 / 1.013 & 0.624 / 0.615 & 0.776 \\
\, \, 1 in one     & 0.244 & 0.251 & 0.279 & 0.091 / 0.091 & 1.023 / 1.016 & 0.574 / 0.563 & 0.090 / 0.093 & 1.020 / 1.021 & 0.612 / 0.593 & 0.792 \\
\, \, avg          & 0.237 & 0.246 & 0.271 & 0.089 / 0.089 & 1.016 / 1.010 & 0.584 / 0.566 & 0.088 / 0.090 & 1.011 / 1.012 & 0.621 / 0.607 & 0.761 
\\
\bottomrule
\end{tabular}}
\caption{Results on 100 validated NLI instances of explanation selection strategy in individual runs. }\label{tab:results-validation-raw}
\end{table*}

\begin{table*}[t]
\centering
\resizebox{\textwidth}{!}{
\begin{tabular}{lccc|ccc|cc|c}
\toprule
\multicolumn{1}{c}{\multirow{2}{*}{\textbf{Datasets}}} & \multicolumn{3}{c|}{\textbf{Lexical}} & \multicolumn{3}{c|}{\textbf{Syntactic}} & \multicolumn{2}{c|}{\textbf{Semantic}} &  \textbf{AVG}\\ \cmidrule(lr){2-10} 
\multicolumn{1}{c}{}          & {n = 1 $\downarrow$} & {n = 2 $\downarrow$} & {n = 3 $\downarrow$}                                     & {n = 1 $\downarrow$} & {n = 2 $\downarrow$} & {n = 3 $\downarrow$}         & {Cos. $\downarrow$}  &  {Euc. $\downarrow$}  & {AVG $\downarrow$}    \\
\midrule
human explanations & 0.339 & 0.103 & 0.045 & 0.753 & 0.340 & 0.140 & 0.512 & 0.516 & 0.343 \\
\midrule
\multicolumn{10}{l}{{first model explanations}}   \\
\, \, Label-Free   & 0.465 & 0.188 & 0.105 & 0.878 & 0.482 & 0.229 & 0.599 & 0.543 & 0.436 \\
\, \, VariErr Label-Guided  & 0.456 & 0.170 & 0.083 & 0.897 & 0.510 & 0.241 & 0.584 & 0.538 & 0.435 \\
\, \, MNLI Label-Guided   & 0.431 & 0.147 & 0.066 & 0.890 & 0.487 & 0.215 & 0.567 & 0.531 & 0.417 \\
\multicolumn{10}{l}{{longest model explanations}}   \\
\, \, Label-Free     & 0.439 & 0.139 & 0.065 & 0.920 & 0.520 & 0.227 & 0.559 & 0.527 & 0.425 \\
\, \, VariErr Label-Guided   & 0.457 & 0.162 & 0.079 & 0.920 & 0.535 & 0.252 & 0.569 & 0.532 & 0.438 \\
\, \, MNLI Label-Guided     & 0.437 & 0.141 & 0.064 & 0.917 & 0.523 & 0.235 & 0.549 & 0.525 & 0.424
\\
\bottomrule
\end{tabular}}
\caption{Linguistic variability check for the main results in Table~\ref{tab:results-main}. }\label{tab:results-similarity-centric-main}
\end{table*}

\section{ANLI Data Contamination Discussion}
\label{app:data-contamination}

\paragraph{Categorizing ANLI as \emph{Out-of-Distribution}}

While the ANLI dataset was indeed released in 2019 and falls within the knowledge cutoff of many LLMs, with the term \emph{out-of-distribution} in this context we refer to the nature of the dataset's construction and its adversarial challenge rather than its temporal release. ANLI was specifically designed to test models on adversarial examples that are difficult to generalize to, even if the models have seen similar data during training. These examples are crafted to exploit weaknesses in reasoning and inference, making them inherently challenging and \emph{out-of-distribution} in terms of difficulty, even if the underlying topics or knowledge are within the model's training scope. For example, in the 2024 CONDA Shared Task~\cite{sainz-etal-2024-data}, ANLI is still used as one of the evaluation benchmarks for current LLMs, which demonstrates the difficulty of ANLI.

\paragraph{Suboptimal Performance and Data Contamination}

The suboptimal performance of LLMs on ANLI does not conclusively rule out data contamination, but it does suggest that the dataset poses a significant challenge even if some contamination exists. ANLI's adversarial nature means that even if some examples were seen during training, the models would still struggle due to the dataset's design. In fact, \citet{DBLP:conf/nips/ZhaYLH23} and \citet{madaan2024lost} have already presented cases of instance overlap between LLM training data and ANLI. Their findings indicate a high overlap ratio. However, both studies argue that this high overlap does not necessarily imply data contamination, as ANLI’s more difficult labels have not been leaked.

To further clarify, we acknowledge that data contamination is a possibility, but the primary focus of our analysis is on the dataset's adversarial difficulty, which remains a robust test of generalization regardless of contamination. In summary, ANLI is considered \emph{out-of-distribution} due to its adversarial construction, and while data contamination cannot be entirely ruled out, the dataset's design ensures it remains a challenging benchmark for evaluating model performance.

\end{document}